%% file: main.tex
\documentclass{article} 
\usepackage{iclr2027_conference,times}

\input{math_commands.tex}

\usepackage{float}
\usepackage{booktabs}
\usepackage{array}
\usepackage{tikz}
\usepackage{hyperref}
\usepackage{xurl}
\DeclareUrlCommand{\identifier}{\urlstyle{rm}}
\defcitealias{InteriorGS2025}{Spatialverse Research Team}
\newcommand{\citeinteriorgs}{(\citetalias{InteriorGS2025}, \citeyear{InteriorGS2025})}
\input{review_changes}

\newcommand{\cerebellum}{Learned Executor}

\title{SuperNav: An Agentic Navigation System\\for Any Task in Any Scene}

\author{%
Jinkai Zhang$^{1}$ \quad Jingyi Xu$^{1}$ \quad Yuanhong Yu$^{1}$ \quad Jiarui Guo$^{1}$ \\
\bfseries Ruizhen Hu$^{2}$ \quad Hujun Bao$^{1}$ \quad Xiaowei Zhou$^{1}$ \quad Sida Peng$^{1,3\dagger}$ \\[4pt]
\normalfont $^{1}$Zhejiang University \quad $^{2}$Shenzhen University \quad $^{3}$\review{Logic Origin} \\
\normalfont $^{\dagger}$Corresponding author
}

\hypersetup{
  pdftitle={SuperNav: An Agentic Navigation System for Any Task in Any Scene},
  pdfauthor={Jinkai Zhang, Jingyi Xu, Yuanhong Yu, Jiarui Guo, Ruizhen Hu, Hujun Bao, Xiaowei Zhou, Sida Peng},
  pdfsubject={Embodied navigation},
  pdfkeywords={SuperNav, embodied navigation, multimodal agents},
  hidelinks
}

\definecolor{lightpink}{RGB}{233, 23, 115}
\definecolor{first}{RGB}{220, 245, 220}   
\definecolor{second}{RGB}{255, 255, 200}  
\definecolor{third}{RGB}{200, 230, 255}   

\iclrfinalcopy 
\begin{document}

\maketitle
\lhead{Preprint}
\vspace{-30pt}

\begin{figure}[H]
\centering
\includegraphics[width=0.85\linewidth]{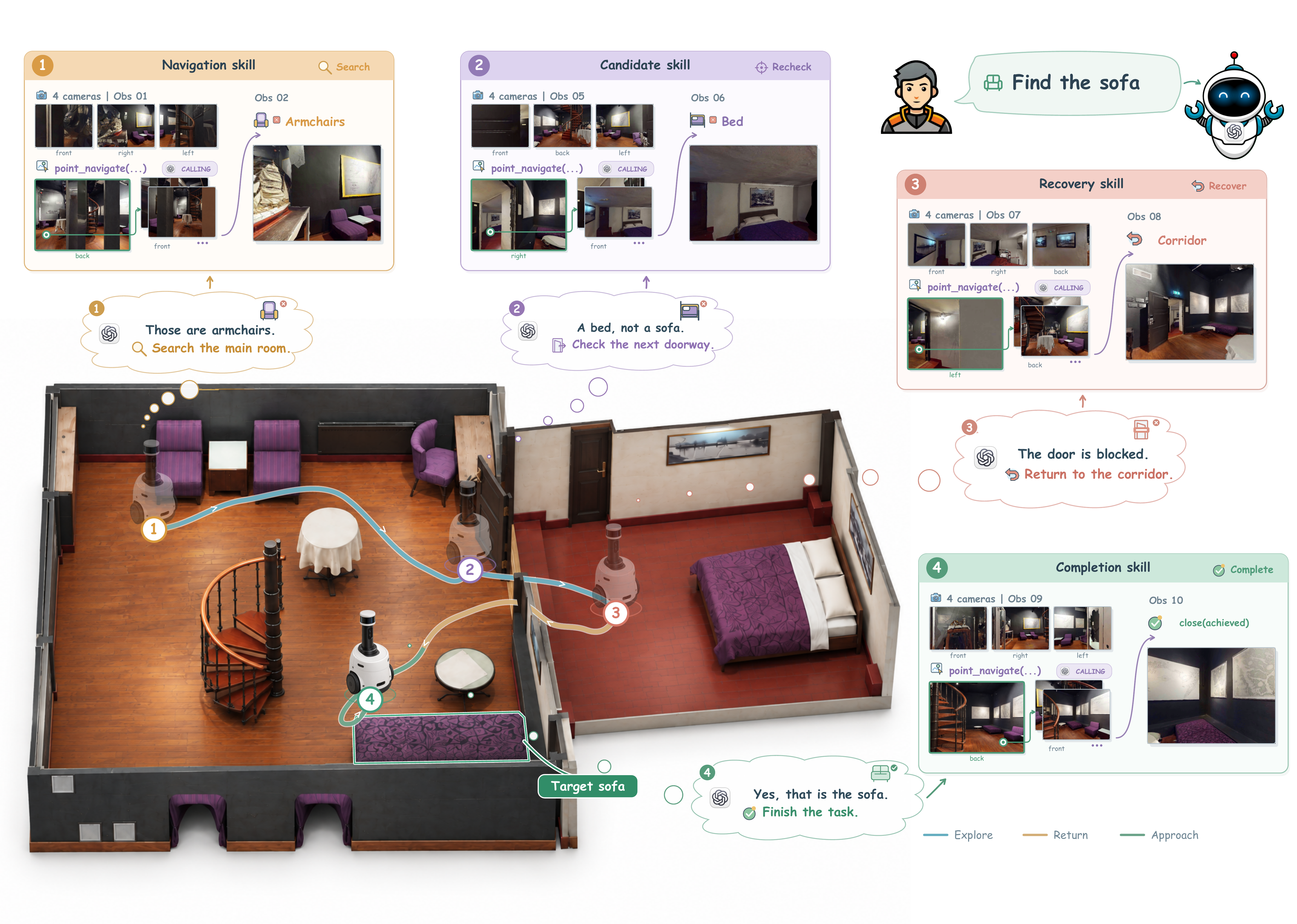}
\caption{\textbf{Skill-guided sofa search.} The Agent rechecks candidates, recovers from a blocked doorway, and verifies completion.}
\label{fig:teaser}
\end{figure}
\vspace{-12pt}

\begin{abstract}
General-purpose service robots need navigation systems that can handle diverse human requests in unfamiliar environments, combining task generality with scene generality.
Some existing methods fine-tune multimodal large language models (MLLMs) to predict navigation actions, making their behavior dependent on the coverage of navigation training data and potentially limiting generalization to new requests and environments.
Our key insight is to let the MLLM focus on interpreting requests, understanding scenes, and making decisions while preserving its general-purpose capabilities and delegating motion execution to navigation tools.
To realize this idea, we introduce \textbf{SuperNav}, which equips a pretrained MLLM with a specialized agent harness without navigation-specific fine-tuning of the MLLM.
Our harness supports these decisions with Navigation Skills, agent-oriented Tools for physical interaction, and task-progress and context management.
A unified visual-point interface connects decision-making to motion by allowing the model to specify destinations directly in images and revise its decisions from execution feedback.
Together, these components support sustained navigation across different task requirements and environments.
SuperNav outperforms four evaluated baselines on instance-level, multi-object, and demand-driven tasks.
Category-level evaluation on HM3D and deployment on a real quadruped robot further demonstrate its applicability across environments.
Project Page: \url{https://zju3dv.github.io/SuperNav/}
\end{abstract}

\input{sections/introduction}
\input{sections/related_work}
\input{sections/method}
\input{sections/experiments}
\input{sections/conclusion}

\section*{AI Use Statement}
We used Codex with GPT-6 to assist with drafting and refining manuscript text based on author-developed ideas.
AI tools also helped identify some comparison papers.
Codex assisted with implementing navigation tools and evaluation code, and with running and debugging experiments.
The authors developed the research ideas, Skill and Tool design principles, and interface specifications, and tested the AI-assisted code.
For our instance-navigation benchmark, generative AI assisted in producing scene-based object annotations and evaluation instructions, which were manually checked by the authors before use.
Images and trajectories were generated by the simulator.
The authors take responsibility for the final content of this work.

\bibliography{iclr2027_conference}
\bibliographystyle{iclr2027_conference}

\appendix
\input{sections/method_implementation}
\input{sections/experiment_details}

\input{sections/method_localnav}
\input{sections/realworld_trajectories}

\end{document}

%% file: math_commands.tex
\usepackage{amsmath,amsfonts,bm}

\def\eqref#1{equation~\ref{#1}}

\def\1{\bm{1}}

\DeclareMathAlphabet{\mathsfit}{\encodingdefault}{\sfdefault}{m}{sl}
\SetMathAlphabet{\mathsfit}{bold}{\encodingdefault}{\sfdefault}{bx}{n}



%% file: review_changes.tex
\usepackage{xcolor}
\newcommand{\review}[1]{\textcolor{black}{#1}}
\newenvironment{reviewblock}{\par\begingroup\color{black}}{\par\endgroup}

\makeatletter
\let\review@bibitem\bibitem
\RenewDocumentCommand{\bibitem}{o m}{%
  \color{black}%
  \IfNoValueTF{#1}{\review@bibitem{#2}}{\review@bibitem[#1]{#2}}%
  \color{black}%
}
\makeatother

%% file: sections/introduction.tex

\section{Introduction}
\label{sec:intro}

General-purpose service robots need navigation capabilities that can fulfill diverse human requests in unfamiliar environments.
These requests range from finding a specific object and visiting multiple targets to satisfying high-level needs and gathering evidence for embodied questions~\citep{objectnav,goatbench,wang2023ddn,mtu3d}.
For example, finding a specified object requires identifying the intended target, whereas ``finding a place to rest'' requires interpreting the need and deciding which objects in the scene can satisfy it.
Such differences change not only what the robot seeks, but also how it explores, tracks progress, and determines completion.
General navigation therefore requires both \emph{task generality}, adapting to different requests, and \emph{scene generality}, operating effectively in environments unseen during training and system development.

Task generality remains constrained by predefined workflows in modular zero-shot navigation systems.
These methods typically use pretrained vision-language models to assess the semantic relevance of observations to a goal and incorporate these assessments into goal-directed exploration scores~\citep{vlfm,apexnav}.
Although these scores adapt search to new observations, task-specific logic still determines when the model participates, how exploration transitions to goal confirmation, and when execution ends.
When a request changes from finding an object category to identifying a particular instance, visiting multiple targets, or satisfying an abstract need, its goal interpretation, progress tracking, and completion conditions also change, often requiring corresponding changes to this workflow.

End-to-end navigation methods seek to support diverse tasks within a single model, but their generalization across both tasks and scenes remains constrained by the coverage of navigation training data.
These approaches typically fine-tune pretrained vision-language models on paired navigation instructions and trajectories to predict actions from visual observations and language~\citep{navid,navillm,uninavid}.
The resulting models encode associations between observations, instructions, and navigation behavior; transfer depends on whether those associations remain useful under new visual appearances, spatial layouts, and interaction conditions.
Consequently, learning to navigate in the training environments does not establish reliable navigation beyond them, and our cross-environment evaluations reveal substantial task-completion gaps for the evaluated methods (Section~\ref{sec:exp_tasks}).

To address these constraints, we introduce \textbf{SuperNav}, a navigation framework that equips a pretrained multimodal large language model (MLLM) with a specialized agent harness for task and scene generality (Figure~\ref{fig:teaser}).
Our design is motivated by the success of general-purpose agents across diverse computer-use tasks, where harnesses support sustained tool use and context management~\citep{lopopolo2026harness,young2025harness}.
Specifically, we use the MLLM's existing capabilities to interpret requests, understand visual observations, and reason about subsequent actions.
To turn these decisions into sustained physical interaction, our harness provides optional \emph{Navigation Skills} for procedural guidance and \emph{agent-oriented Tools} for observation, motion, and task management.
It also maintains task-progress records and manages the interaction context, enabling the model to continue reasoning across successive actions.
This division of responsibilities lets the MLLM adapt its use of navigation capabilities to the current request and scene without navigation-specific fine-tuning.

Connecting these decisions to physical motion requires an interface that grounds the MLLM's text outputs in the observed scene.
We therefore design a \emph{unified visual-point interface}: the MLLM specifies a destination directly in an observation image, and the navigation tool executes the movement and returns feedback for the next decision.
Compared with predicting low-level actions directly~\citep{navid,uninavid}, this interface lets the model focus on selecting destinations while supporting different motion backends through the same interaction pattern.
SuperNav achieves higher success rates than these two action-prediction baselines on our single-object and demand-driven task sets (Table~\ref{tab:local_navigation}).

We construct an instance-navigation benchmark in Habitat-GS~\review{\citep{xia2026habitat}} using InteriorGS scenes~\review{\citeinteriorgs} to evaluate single-object and ordered multi-object navigation.
We also evaluate demand-driven navigation using tasks from Demand-Bench~\citep{anonymous2026demandagent} under an adapted protocol, and category-level navigation on HM3Dv2 and HM3D-OVON, complemented by real-quadruped deployment.
SuperNav achieves 78.00\% success on single-object navigation and 59.50\% on demand-driven navigation, compared with 34.00\% and 37.50\% for the respective strongest baselines, UniNaVid and OmniNav's Action Former (Table~\ref{tab:local_navigation}).
For category-level navigation, SuperNav achieves 73.33\% success rate (SR) and 0.4105 success weighted by path length (SPL) on a 120-episode HM3D-OVON val-unseen subset at a 1\,m threshold, without OVON-specific training or adaptation.
At the stricter 0.25\,m threshold, it achieves 68.33\% SR and 0.3837 SPL (Table~\ref{tab:hm3d_results}).
These results examine the complete system across task requirements and scene settings, with motion-backend effects and evaluation-protocol differences analyzed in Section~\ref{sec:experiments}.

Our contributions are threefold:
\begin{itemize}
    \item We introduce SuperNav, a specialized agent harness that enables a pretrained MLLM to direct navigation across different requests through tool execution, task-progress tracking, and context management.
    \item We design agent-oriented Tools with a unified visual-point motion interface and optional Navigation Skills, connecting model decisions to physical execution and providing reusable guidance for search, verification, and recovery.
    \item We evaluate the system across category-level, instance-level, multi-object, and demand-driven navigation, together with real-robot deployment, examining its applicability across task and scene settings.
\end{itemize}

%% file: sections/related_work.tex

\section{Related Work}
\label{sec:related}

\paragraph{Modular Navigation Pipelines.}
Modular systems integrate semantic reasoning into dedicated mapping, exploration, and verification components.
SemExp~\citep{semexp}, L3MVN~\citep{l3mvn}, and VLFM~\citep{vlfm} guide object search through learned exploration, language priors, or image--text relevance.
VoroNav~\citep{voronav} and SG-Nav~\citep{sgnav} add topological or scene-graph context, while ApexNav~\citep{apexnav} combines adaptive exploration with multi-frame target identification.
These systems assign models specific roles within task-oriented pipelines; SuperNav instead lets the MLLM select and coordinate available capabilities at runtime.

\paragraph{Hierarchical Navigation Systems.}
Hierarchical systems couple high-level model reasoning with dedicated navigation modules.
InstructNav~\citep{instructnav} translates predicted actions and landmarks into trajectories through value maps; OmniNav~\citep{omninav} combines prospective exploration and subgoal planning with a fast waypoint policy; SysNav~\citep{sysnav} delegates within-room exploration and motion to classical algorithms under room-level VLM reasoning.
These systems already support online adaptation, demonstrating cooperation between model reasoning and learned or geometric execution.
SuperNav focuses on the interfaces and runtime through which the model chooses and combines capabilities across a task.

\paragraph{Agentic Systems and Harnesses.}
Agentic systems organize execution through callable capabilities and feedback.
Inner Monologue~\citep{inner_monologue} incorporates environmental feedback into planning, Code as Policies~\citep{code_as_policies} composes robotics APIs into executable policies, and SWE-agent~\citep{sweagent} studies how action and observation interfaces affect software-task performance.
Embodied harnesses extend this perspective: Show-Harness~\citep{showharness} translates semantic actions into robot motion, AgenticNav~\citep{agenticnav} combines pixel-goal motion with depth queries and historical-image retrieval, and HarnessVLN~\citep{harnessvln} maintains event memory and a spatiotemporal graph while validating action and termination proposals.
Concurrent DemandAgent~\citep{anonymous2026demandagent} addresses dependency-aware human demands through an Adaptive Demand Ledger, spatial memory, and embodied Skills, and introduces Demand-Bench. SuperNav studies task and scene generality through a unified visual-point interface with interchangeable motion backends, supported by context pruning and Navigation Skills.
Task-level action selection and semantic completion judgments remain with the MLLM.

%% file: sections/method.tex
\section{Navigation Agent Harness}
\label{sec:method}

\paragraph{Overview.}
SuperNav combines a pretrained MLLM with a navigation-specific harness to form an embodied navigation Agent that sustains decisions and actions across different requests.
A request may specify a target, an ordered sequence of targets, or a high-level need; the Agent must interpret that request and act until it can judge whether the completion conditions are met.
The MLLM performs this interpretation and decision-making, while the harness organizes execution and context through an \emph{agent loop}.
Within this loop, \emph{agent-oriented Tools} are callable operations for observation, motion, and task-state management, and \emph{Navigation Skills} are instructions the MLLM can read for search, verification, and recovery.
As illustrated in Figure~\ref{fig:system}, the MLLM interprets observations, selects a Tool call, and uses the returned feedback to decide again, consulting Skills when relevant.
We first describe how the loop coordinates these capabilities (Section~\ref{sec:agent}), then develop the Tool interfaces (Section~\ref{sec:tools}) and Skill guidance (Section~\ref{sec:skills}).

\begin{figure}[!t]
\centering
\input{figures/figure1}
\caption{\textbf{Navigation with Skills and agent-oriented Tools.} \textbf{(a)} The Agent loop maintains context for MLLM reasoning, Skill loading, and Tool calls. \textbf{(b)} Navigation Skills guide search, candidate rechecking, recovery, and completion; the recovery example shows the Agent retracing its path after a blocked passage. \textbf{(c)} Agent-oriented Tools support observation, motion, goal-progress tracking, and task closure. Point navigation shares an image-space goal interface across geometric and learned execution backends.}
\label{fig:system}
\end{figure}

\subsection{Agent Loop}
\label{sec:agent}

\paragraph{Continuous decision-making.}
The agent loop organizes the request, visual observations, and interaction history into a continuing navigation process.
Initialization opens a session and returns the first observation, from which the MLLM interprets the requested goals and completion conditions.
At each subsequent step, it uses the current context and task progress to choose an operation: observe, move, update goal state, or read relevant Skill guidance.
Tool returns and retrieved instructions enter the context for the next decision, allowing execution feedback to change the MLLM's plan.
For example, after approaching a candidate, the MLLM inspects the new images, judges whether the candidate satisfies the request, and either records completion or continues searching.

\paragraph{Task state and context management.}
Long-horizon execution requires these local decisions to remain connected to earlier evidence and unfinished goals.
Goal records maintain pending and completed items, while textual interaction history retains inspected areas, candidate clues, and failed attempts for later decisions.
To control the growing image context, the harness removes an image's media payload from subsequent requests once it has been processed and superseded by newer observations.
It preserves textual history, task state, and image paths; the MLLM can request earlier images from the workspace when needed.
This media-only pruning limits repeated image transmission while retaining access to earlier evidence and task progress.

\paragraph{Task termination.}
The MLLM judges whether the request is satisfied or execution is blocked from the observations and progress records, and selects the termination Tool to end the loop.
A termination Tool records either declared outcome and closes the session; successful closure confirms that the session ended, not that the task succeeded.
Benchmark scoring evaluates success separately from this declaration.
The loop therefore depends on operations that expose observations, execute motion, and maintain state, whose interfaces we describe next; runtime restrictions and execution limits appear in Appendix~\ref{sec:impl}.

\subsection{Agent-oriented Navigation Tools}
\label{sec:tools}

\paragraph{Tool suite and design principles.}
Agent-oriented Tools connect the MLLM's navigation decisions to executable environment operations.
The suite covers initialization, observation, turning, visual-point navigation, goal-progress management, and termination.
Its design makes inputs easy for the MLLM to specify and returns useful for the next decision, while keeping destination representations consistent across motion backends.
The central observation--motion interface implements these principles by pairing visual destinations with updated observations and execution feedback.

\paragraph{Observation and visual-point navigation.}
An observation returns four RGB images labeled \emph{front}, \emph{right}, \emph{back}, and \emph{left}, relative to the robot's heading at capture.
These labels let the MLLM associate a candidate or passage with a particular view and select a destination through
\[
\mathrm{PointNav}(d,[u,v]), \qquad u,v\in[0,1],
\]
where $d$ identifies the view and $u,v$ are horizontal and vertical coordinates normalized from its top-left corner.
The Agent may select a point on a candidate or on visible floor to approach a target or obtain a new view.
The Tool executes the movement internally and returns new four-view images, execution status, and concise diagnostics; the turning Tool provides the same types of feedback.
These returns connect destination selection to visual rechecking without requiring a separate observation call after every movement.

\paragraph{Interchangeable motion backends.}
The visual-point interface separates destination selection from the mechanism that executes it.
The geometric backend, termed \emph{Geo-based Executor}, uses depth, camera calibration, and pose to associate the selected point with an environment location, then plans and follows a path to a reachable goal.
Simulation supplies rendered depth, camera poses, and navigation geometry, while robot deployment uses sensed geometry and localization (Appendix~\ref{sec:experimental_setup}).
The learned backend, \cerebellum{}, is inspired by NoMaD's goal-conditioned generative-policy design~\citep{nomad}, replacing destination-image goals with marked points in reference observations.
It predicts local displacement increments and a stopping signal from the selected image--point pair and recent RGB observations.
It turns toward the selected view and repeatedly executes short predicted motion segments within one Tool call.
Both backends accept visual destinations and return observations with execution feedback, so the Agent can retain the same decision workflow; learned-controller architecture and training details are deferred to Appendix~\ref{sec:localnav}.

\paragraph{Progress and termination.}
Progress and termination Tools extend this workflow from individual movements to complete requests.
The progress Tool lets the MLLM query or update \emph{pending}/\emph{found} goals and continue with the remaining items; the termination Tool records its \emph{achieved}/\emph{blocked} judgment and closes the session.
These records represent the MLLM's reports rather than independent semantic verification.
Deciding when to search further, recheck a candidate, or recover from failed motion thus remains the MLLM's responsibility, for which Skills provide reusable guidance.

\subsection{Navigation Skills}
\label{sec:skills}

\paragraph{Skill organization and discovery.}
Navigation Skills organize reusable navigation experience into procedural guidance that the MLLM can read on demand.
Each Skill is a Markdown package describing its applicable situations, evidence to inspect, recommended steps, and linked references.
The bundle comprises a main navigation Skill, backend-specific Tool-use instructions, and references for exploration, recovery, and entrance search, together covering search, recovery, and completion.
The harness exposes Skill names and descriptions so the MLLM can discover the relevant package and load its instructions or linked references into the decision context.

\paragraph{Navigation guidance.}
The guidance addresses three recurring decisions: where to search, how to recover, and when to finish.
\textbf{Search guidance} asks the MLLM to record inspected areas, visible openings, and uncertain branches, preserving alternatives for later exploration; revisiting relies on visual recognition rather than a stored return path.
\textbf{Recovery guidance} recommends changing the point or route after failed motion, stepping back when a candidate is difficult to inspect, or seeking an entrance when glass blocks access.
\textbf{Completion guidance} calls for checking the candidate's appearance, requested relations, and surrounding context, then recording verified goals and continuing with any remaining request.
For example, a blocked passage can prompt the MLLM to read recovery guidance, select another route, and reassess the candidate from the returned images.

\paragraph{Integration with the Agent loop.}
Together, the loop sustains decisions, Tools provide execution capabilities, and Skills guide their use.
Skills influence decisions by entering the context; they neither add environment capabilities nor programmatically enforce action sequences or semantic judgments.
The MLLM still chooses each operation from the current request and feedback, using the shared loop, interfaces, and guidance to handle different navigation tasks.

%% file: figures/figure1.tex
\includegraphics[width=\linewidth]{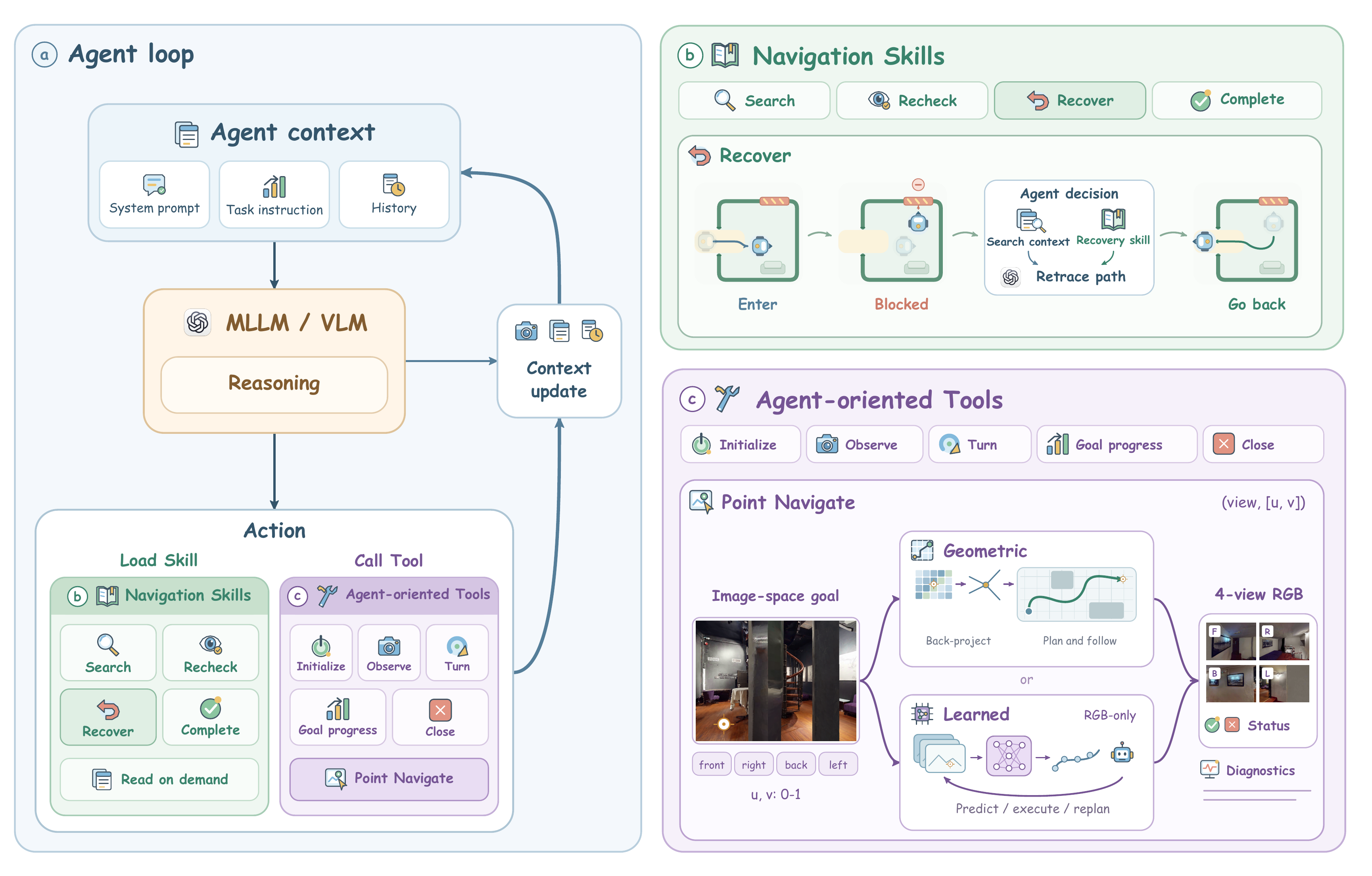}

%% file: sections/experiments.tex
\section{Experiments}
\label{sec:experiments}
\label{sec:exp_setup}

We evaluate SuperNav across four settings using GPT-5.6 Terra with Geo-based Executor or Learned Executor, reporting SR (\%) and SPL (0--1).
In Habitat-GS~\review{\citep{xia2026habitat}} and AI2-THOR, we compare against NaVid~\citep{navid}, UniNaVid~\citep{uninavid}, StreamVLN~\citep{streamvln}, and OmniNav's Action Former~\citep{omninav}, retaining each baseline's native inputs and history mechanisms.
Detailed protocols appear in Appendix~\ref{sec:exp_details}.

\subsection{Instance Navigation in Habitat-GS}
\label{sec:exp_tasks}

\review{Our benchmark, based on InteriorGS~\citeinteriorgs,} contains 150 single-object tasks and 150 ordered multi-object tasks with two to five targets.
Arrival requires navmesh geodesic distance below 1\,m to an approved target viewpoint.
Single-object success tests the final position; multi-object success tests ordered arrivals along the trajectory, both requiring explicit STOP (Appendix~\ref{sec:gs_construction}--\ref{sec:exp_progress}).

\begin{table}[!t]
\centering
\caption{\textbf{Single-object, multi-object, and demand-driven navigation results.} SR (\%) and SPL (0--1): instance tasks use viewpoint geodesic distance $<1$\,m; demand-driven tasks use 2\,m bounding-box regions. Task counts are in parentheses; bold marks column bests.}
\label{tab:local_navigation}
\small
\setlength{\tabcolsep}{4pt}
\begin{tabular}{@{}lrrrrrr@{}}
\toprule
& \multicolumn{4}{c}{Our instance-navigation benchmark} & \multicolumn{2}{c}{Demand-driven benchmark} \\
\cmidrule(lr){2-5}\cmidrule(l){6-7}
& \multicolumn{2}{c}{Single-object (150)} & \multicolumn{2}{c}{Multi-object (150)} & \multicolumn{2}{c}{Demand-driven (200)} \\
\cmidrule(lr){2-3}\cmidrule(lr){4-5}\cmidrule(l){6-7}
Method & SR$\uparrow$ & SPL$\uparrow$ & SR$\uparrow$ & SPL$\uparrow$ & SR$\uparrow$ & SPL$\uparrow$ \\
\midrule
NaVid & 24.67 & 0.1599 & 2.67 & 0.0216 & 17.00 & 0.0920 \\
UniNaVid & 34.00 & 0.1833 & 1.33 & 0.0131 & 25.00 & 0.0943 \\
StreamVLN & 13.33 & 0.1036 & 0.00 & 0.0000 & 25.50 & 0.0775 \\
OmniNav (Action Former) & 27.33 & 0.2133 & 4.00 & 0.0275 & 37.50 & 0.1213 \\
\midrule
SuperNav + Geo-based Executor & \textbf{78.00} & \textbf{0.4127} & \textbf{34.00} & \textbf{0.1388} & \textbf{59.50} & \textbf{0.1801} \\
SuperNav + Learned Executor & 68.00 & 0.2059 & \textbf{34.00} & 0.0862 & 45.00 & 0.1603 \\
\bottomrule
\end{tabular}
\end{table}

\begin{figure}[!t]
\centering
\includegraphics[width=\linewidth]{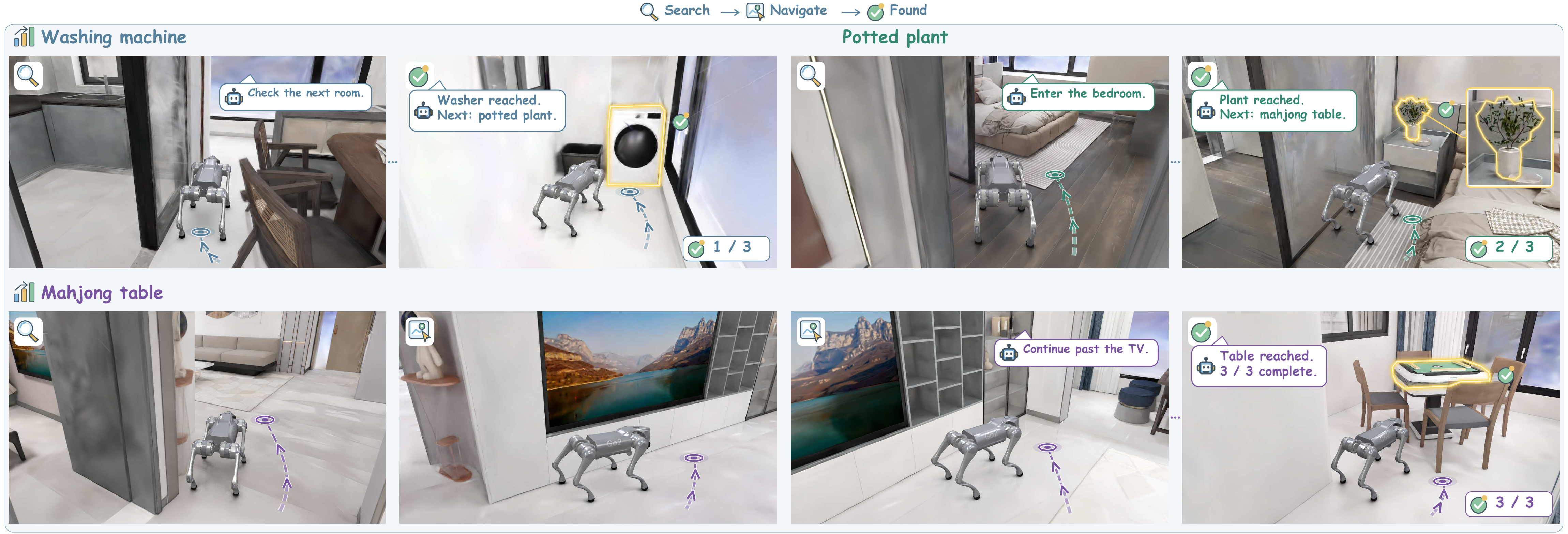}
\caption{\textbf{Ordered multi-object navigation in simulation.} The Agent visits a washing machine, a potted plant, and a mahjong table in order.}
\label{fig:simulation_navigation}
\end{figure}

\paragraph{Results.}
SuperNav sustains its advantage through later goals (Table~\ref{tab:local_navigation}; Appendix~\ref{sec:exp_progress}), as illustrated in Figure~\ref{fig:simulation_navigation}.
Both backends achieve the same multi-object success rate, while Geo-based Executor yields more efficient paths.

\subsection{Demand-Driven Navigation}
\label{sec:exp_demand}

We use 200 AI2-THOR~\citep{ai2thor} tasks from Demand-Bench~\citep{anonymous2026demandagent}, adapting evaluation to one to eight ordered stages (Appendix~\ref{sec:local_metrics}).
Success requires entering each stage's admissible 2\,m bounding-box region in order, without object reuse, and explicitly stopping.

\paragraph{Results.}
Both backends improve on the baselines in SR and SPL (Table~\ref{tab:local_navigation}), extending the framework from explicit targets to inferred goals.
Geo-based Executor progresses furthest, but long sequences remain challenging (Appendix~\ref{sec:exp_progress}).

\begin{figure}[!t]
\centering
\includegraphics[width=\linewidth]{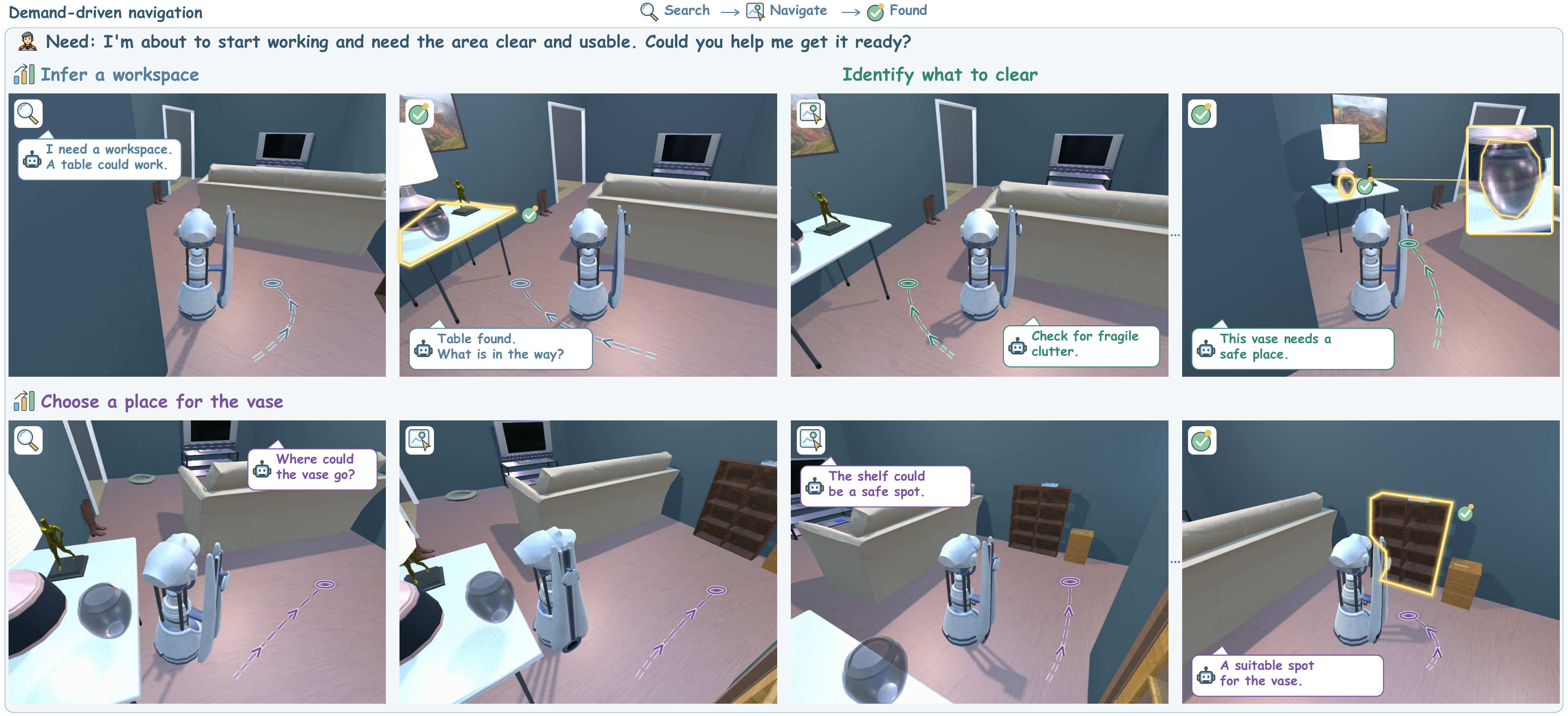}
\caption{\textbf{Demand-driven navigation.} To prepare a workspace, the Agent locates a table, identifies a vase to clear, and finds a suitable shelf for it.}
\label{fig:demand_driven_navigation}
\end{figure}

Figure~\ref{fig:demand_driven_navigation} illustrates navigation toward successive inferred goals.

\subsection{HM3D Benchmarks}
\label{sec:exp_hm3d}

We evaluate category-level navigation on a shared 120-episode HM3D-OVON val-unseen subset and 1,000 HM3Dv2 validation episodes~\citep{hm3d,ovon}.
\review{Table~\ref{tab:hm3d_results} groups results by distance threshold, using Geo-based Executor or Learned Executor.}
Our scoring rules and the protocols behind published comparisons are detailed in Appendix~\ref{sec:hm3d_protocol}--\ref{sec:ovon_ablation_details}.

\begin{table}[!t]
\centering
\caption{\textbf{Category-level navigation on HM3D.} SR (\%) and SPL (0--1) at the indicated thresholds. \review{HarnessVLN and InstructNav-adapted are evaluated on our shared 120-episode OVON subset. Other external} entries are published results; L3MVN, VLFM, and SG-Nav use ApexNav's HM3Dv2 reevaluation. \review{The public simulation implementations of L3MVN, VLFM, SG-Nav, ApexNav, MSGNav, and HarnessVLN also use privileged information: GT depth and GT pose for online mapping. WMNav's default simulation configuration likewise uses GT depth and GT pose. VLFM's PointNav executor also uses GT depth and relative goal distance and heading computed from GT pose.} \review{GGF: Greedy Geodesic Follower; FMM: Fast Marching Method.} Bold marks the highest value in each column.}
\label{tab:hm3d_results}
\small
\setlength{\tabcolsep}{2.5pt}
\resizebox{\linewidth}{!}{%
\begin{tabular}{@{}llrrrrrrrr@{}}
\toprule
& & \multicolumn{4}{c}{HM3D-OVON val-unseen} & \multicolumn{4}{c}{HM3Dv2} \\
\cmidrule(lr){3-6}\cmidrule(l){7-10}
& & \multicolumn{2}{c}{0.25\,m} & \multicolumn{2}{c}{1\,m} & \multicolumn{2}{c}{0.2\,m} & \multicolumn{2}{c}{1\,m} \\
\cmidrule(lr){3-4}\cmidrule(lr){5-6}\cmidrule(lr){7-8}\cmidrule(l){9-10}
Method & \review{Motion executor} & SR$\uparrow$ & SPL$\uparrow$ & SR$\uparrow$ & SPL$\uparrow$ & SR$\uparrow$ & SPL$\uparrow$ & SR$\uparrow$ & SPL$\uparrow$ \\
\midrule
MTU3D~\citep{mtu3d} & \review{GT navmesh + GGF} & 40.80 & 0.1210 & --- & --- & --- & --- & --- & --- \\
SoftNav~\citep{softnav} & \review{Unverified} & 66.70 & 0.2570 & --- & --- & --- & --- & --- & --- \\
AstraNav-Memory~\citep{astranav} & \review{GT navmesh + GGF} & --- & --- & 62.50 & 0.3490 & --- & --- & --- & --- \\
OmniNav (slow + CoT)~\citep{omninav} & \review{GT navmesh + GGF} & --- & --- & 59.20 & 0.3320 & --- & --- & --- & --- \\
L3MVN~\citep{l3mvn} & \review{Online map + FMM} & --- & --- & --- & --- & 36.30 & 0.1570 & --- & --- \\
VLFM~\citep{vlfm} & \review{Learned PointNav} & --- & --- & --- & --- & 63.60 & 0.3250 & --- & --- \\
SG-Nav~\citep{sgnav} & \review{Online map + FMM} & --- & --- & --- & --- & 49.60 & 0.2550 & --- & --- \\
ApexNav~\citep{apexnav} & \review{Online map + A*} & --- & --- & --- & --- & 76.20 & \textbf{0.3800} & --- & --- \\
WMNav~\citep{wmnav} & \review{Polar motion primitives} & --- & --- & --- & --- & --- & --- & 72.20 & 0.3330 \\
MSGNav~\citep{msgnav} & \review{GT navmesh + waypoint teleportation} & --- & --- & --- & --- & --- & --- & 74.10 & 0.3340 \\
\review{HarnessVLN~\citep{harnessvln}} & \review{Map/FMM + NavDP~\citep{cai2025navdp}} & \review{---} & \review{---} & \review{53.33} & \review{0.2941} & \review{---} & \review{---} & \review{---} & \review{---} \\
\review{InstructNav-adapted~\citep{instructnav}} & \review{GT navmesh + GGF} & \review{41.67} & \review{0.1865} & \review{54.17} & \review{0.2306} & \review{---} & \review{---} & \review{---} & \review{---} \\
\midrule
\review{SuperNav} & \review{GT navmesh + GGF} & \textbf{68.33} & \textbf{0.3837} & \textbf{73.33} & \textbf{0.4105} & \textbf{80.30} & 0.3338 & \textbf{86.50} & \textbf{0.3638} \\
\review{SuperNav} & \review{Learned Executor} & 65.00 & 0.1355 & 70.83 & 0.1443 & 79.00 & 0.1288 & 85.40 & 0.1383 \\
\bottomrule
\end{tabular}
}
\end{table}

\paragraph{Results.}
Learned Executor retains much of Geo-based Executor's success but loses path efficiency (Table~\ref{tab:hm3d_results}).
\review{Appendix~\ref{sec:motion_path_efficiency} details the map/planner audit and analyzes path efficiency on episodes successful under both backends.}
Geo-based Executor is competitive with the listed published methods, although their evaluation subsets and protocols differ from ours.
\review{On the shared OVON subset, InstructNav-adapted uses a fixed decision sequence with the same GPT model, four cameras, motion planner, and time budget as SuperNav, achieving 41.67\% and 54.17\% SR at 0.25\,m and 1\,m, respectively (Appendix~\ref{sec:instructnav_adapted}).}

\subsection{Harness and Model Ablations}
\label{sec:exp_ablation}

\begin{reviewblock}
\subsubsection{Navigation Skills, Tool Design, and Decision Model Ablation}
\label{sec:ovon_ablations}
\end{reviewblock}

With Geo-based Executor fixed, we remove Navigation Skills, replace agent-oriented Tools with front-only interaction, or substitute language grounding and candidate selection for direct pointing (\hyperref[par:grounding_motion_tool]{Grounding Motion Tool}, Appendix~\ref{sec:ovon_ablation_details}).
A decision-model configuration variant uses GPT-6 Astra (medium) instead of GPT-5.6 Terra (high), preserving the full harness.

\begin{table}[!t]
\centering
\caption{\textbf{OVON ablations with Geo-based Executor.} Each variant uses 120 episodes and GPT-5.6 Terra (high) unless specified. SR (\%) and SPL (0--1); bold marks the best value.}
\label{tab:ovon_ablation}
\small
\setlength{\tabcolsep}{6pt}
\begin{tabular}{@{}lrrrr@{}}
\toprule
& \multicolumn{2}{c}{0.25\,m} & \multicolumn{2}{c}{1\,m} \\
\cmidrule(lr){2-3}\cmidrule(l){4-5}
Variant & SR$\uparrow$ & SPL$\uparrow$ & SR$\uparrow$ & SPL$\uparrow$ \\
\midrule
w/o Navigation Skills & 35.00 & 0.2460 & 54.17 & 0.3894 \\
w/o Agent-oriented Tool Design & 52.50 & 0.2230 & 65.83 & 0.2855 \\
w/o Skills and Tool Design & 31.67 & 0.1870 & 54.17 & 0.3417 \\
Grounding Motion Tool & 55.83 & 0.3201 & 62.50 & 0.3624 \\
\midrule
SuperNav (GPT-6 Astra medium) & \textbf{71.67} & \textbf{0.4579} & \textbf{79.17} & \textbf{0.4974} \\
SuperNav (GPT-5.6 Terra high) & 68.33 & 0.3837 & 73.33 & 0.4105 \\
\bottomrule
\end{tabular}
\end{table}

\paragraph{Results.}
Removing Navigation Skills reduces SR more than restricting the interface to the front view, especially at the stricter threshold (Table~\ref{tab:ovon_ablation}).
The Tool design complements this guidance: with Skills retained, front-only interaction reduces SR and SPL while requiring substantially more explicit turns (Table~\ref{tab:ovon_ablation_execution}).
External grounding does not improve on the Agent's direct point selection, whereas replacing Terra (high) with Astra (medium) improves both SR and SPL with the harness fixed. This demonstrates compatibility with a different decision-model configuration.
\begin{reviewblock}
\subsubsection{Automatically Generated Navigation Guidance}
\label{sec:generated_guidance}

GPT-6 Astra generates Navigation Skills and a prompt in a fresh context from navigation requirements, tool specifications, environment constraints, and the \identifier{SKILL.md} format.
We evaluate this guidance with GPT-5.6 Terra and Geo-based Executor on the same 150 single-object episodes, holding observations, execution, the 3,600\,s budget, and scoring fixed.

\begin{table}[!t]
\color{black}
\centering
\caption{\textbf{Automatically generated navigation guidance on single-object navigation.} Both configurations use GPT-5.6 Terra and Geo-based Executor on the same 150 episodes. SR (\%) and SPL (0--1) use the 1\,m viewpoint criterion; bold marks the best value.}
\label{tab:generated_guidance}
\small
\setlength{\tabcolsep}{6pt}
\begin{tabular}{@{}lrr@{}}
\toprule
Navigation guidance & SR$\uparrow$ & SPL$\uparrow$ \\
\midrule
Original Skills and prompt & \textbf{78.00} & 0.4127 \\
Automatically generated Skills and prompt & 71.33 & \textbf{0.4542} \\
\bottomrule
\end{tabular}
\end{table}

The generated guidance achieves 71.33\% SR and 0.4542 SPL, compared with 78.00\% and 0.4127 for the original guidance (Table~\ref{tab:generated_guidance}), demonstrating effective navigation with automatically generated Skills and a prompt.

\subsubsection{Progress Management Tool}
\label{sec:progress_ablation}

We remove only the progress management Tool and retain all other settings, using GPT-5.6 Terra and Geo-based Executor on the same 150 multi-object episodes.
Under the viewpoint geodesic distance $<1$\,m criterion, SR decreases from 34.00\% to 31.33\% and SPL from 0.1388 to 0.1198 (Table~\ref{tab:progress_ablation}).

\begin{table}[!t]
\color{black}
\centering
\caption{\textbf{Progress management ablation on multi-object navigation.} Both configurations use GPT-5.6 Terra and Geo-based Executor on the same 150 episodes, with viewpoint geodesic distance $<1$\,m. SR (\%) and SPL (0--1); bold marks the best value.}
\label{tab:progress_ablation}
\small
\setlength{\tabcolsep}{6pt}
\begin{tabular}{@{}lrr@{}}
\toprule
Variant & SR$\uparrow$ & SPL$\uparrow$ \\
\midrule
SuperNav (Full) & \textbf{34.00} & \textbf{0.1388} \\
w/o Progress Management Tool & 31.33 & 0.1198 \\
\bottomrule
\end{tabular}
\end{table}
\end{reviewblock}

\subsection{Real-World Navigation}
\label{sec:exp_realworld}

\begin{figure}[!t]
\centering
\includegraphics[width=\linewidth]{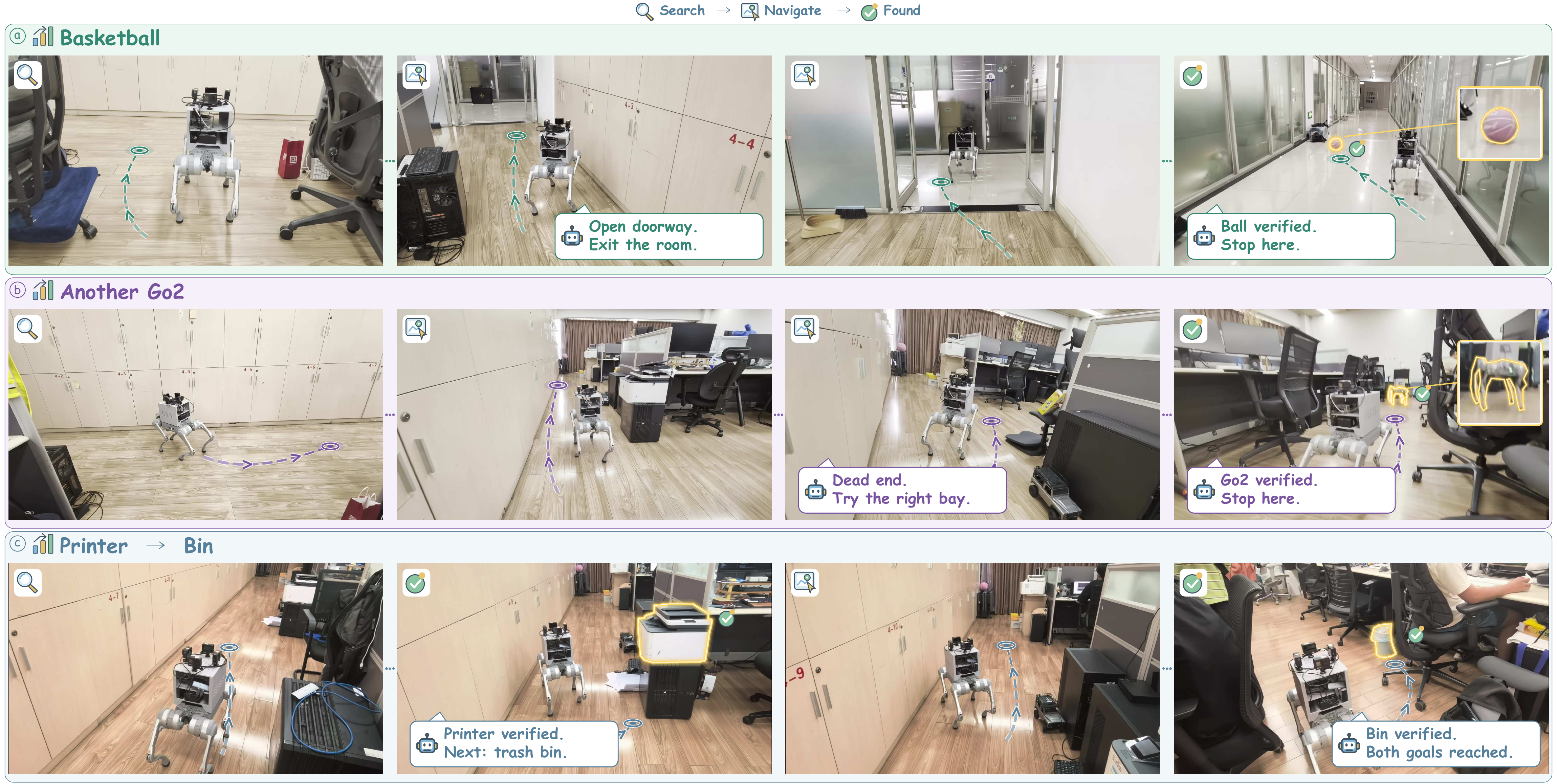}
\caption{\textbf{Real-world navigation on Unitree Go2.} The robot searches for (a) a basketball and (b) another Go2, and (c) visits a printer followed by a trash bin.}
\label{fig:realworld_navigation}
\end{figure}

On a Unitree Go2 with four-view RGB and LiDAR-based geometric execution, SuperNav revises routes at blocked passages, verifies targets, and transitions between goals (Figure~\ref{fig:realworld_navigation}; Appendix~\ref{sec:experimental_setup}).

\begin{reviewblock}
On ten real-world navigation tasks, SuperNav succeeds on nine, while ApexNav and LightNav-0 each succeed on three (Table~\ref{tab:realworld_results}). Appendix~\ref{sec:realworld_task_results} lists the instructions and per-task outcomes; Figures~\ref{fig:realworld_tasks_1_2}--\ref{fig:realworld_tasks_9_10} show the trajectories.

\begin{table}[!t]
\color{black}
\centering
\caption{\textbf{Real-world navigation results on ten tasks.} SR denotes success rate.}
\label{tab:realworld_results}
\small
\begin{tabular}{@{}lrr@{}}
\toprule
Method & Successful tasks & SR (\%)$\uparrow$ \\
\midrule
ApexNav~\citep{apexnav} & 3/10 & 30.0 \\
LightNav-0~\citep{lightnav0} & 3/10 & 30.0 \\
\midrule
SuperNav & \textbf{9/10} & \textbf{90.0} \\
\bottomrule
\end{tabular}
\end{table}
\end{reviewblock}

%% file: sections/conclusion.tex
\section{Conclusion}
\label{sec:conclusion}

We presented SuperNav, a navigation harness that enables a pretrained MLLM to interpret requests, understand scenes, and make decisions while delegating motion execution to tools, without navigation-specific fine-tuning of the MLLM.
Navigation Skills, task-progress tracking, and context management support sustained interaction, while a unified visual-point interface connects decisions to physical motion.
Experiments across category-level, instance-level, multi-object, and demand-driven navigation demonstrate the framework's applicability across task and scene settings.
These results support building navigation capabilities around a general-purpose MLLM's existing understanding and reasoning.
\paragraph{Limitations.}
SuperNav relies on the underlying MLLM for semantic judgments and is therefore limited by its capabilities. Geometric execution relies on available maps or scene geometry, while MLLM inference incurs computational cost and substantial latency variability, making task duration less predictable. Our demand-driven evaluation measures ordered navigation to relevant targets rather than completion of the underlying activities. Extending evaluation to more real-world settings and improving reliability on long navigation sequences remain future directions.

%% file: sections/method_implementation.tex
\section{Method implementation details}
\label{sec:impl}

\subsection{Experimental Setup}
\label{sec:experimental_setup}

\paragraph{Unitree Go2.}
\review{SuperNav's real-world experiments use a Unitree Go2 Air robot equipped with GoPro HERO8 Black cameras.}
The geometric backend combines four directional RGB cameras with robot-provided LiDAR data and odometry.
Camera intrinsics, mounting extrinsics, and the robot pose associated with the selected observation relate image-space goals to the LiDAR map.
The backend associates the selected pixel's neighborhood with mapped 3D points, selecting a foreground surface and an approach position.
A voxel-grid mapper accumulates LiDAR observations into a height-based 2D traversal-cost map; voxel size and grid resolution both default to 0.05\,m.
\identifier{ReplanningAStarPlanner} computes paths, and a lookahead-based follower uses proportional heading control to generate linear and angular velocity commands at 10\,Hz.
Localization uses the robot's reported odometry, without an additional loop-closing SLAM system in our software.

\paragraph{Simulation and Geo-based Executor.}
Habitat-Sim~\review{\citep{habitat19iccv,szot2021habitat,puig2023habitat3}} provides RGB-D images at $640\times480$ resolution, a $90^{\circ}$ horizontal field of view, and a camera height of 1.25\,m, together with known intrinsics and simulator camera poses.
Geo-based Executor back-projects image-space anchors using rendered depth and checks nearby candidate destinations on the scene's precomputed navigation mesh.
Recast/Detour-based \identifier{PathFinder} supplies path queries; \identifier{GreedyGeodesicFollower} executes discrete forward and turning actions.
Defaults are a 0.7\,m standoff, a 0.75\,m candidate snap-distance limit, a 0.3\,m follower goal radius, and 200 follower steps.
These local execution limits are separate from benchmark success thresholds.

\paragraph{Learned Executor controller.}
\cerebellum{} generates motion from RGB observations and image-space goals without depth-based grounding or navigation-mesh planning.
Learned Executor uses marked reference images, a shared DINOv2 encoder, variable-length visual history, and a flow action head, as described in Section~\ref{sec:localnav}.
Simulator position changes and collision reports support progress monitoring.

\subsection{Runtime and Tool Access}

\paragraph{Agent runtime.}
We implement the Agent loop with Codex CLI and expose environment operations through the Model Context Protocol (MCP).

\paragraph{Task workspaces and permissions.}
For Learned Executor and Geo-based Executor, the harness launches each non-interactive Codex process with a task-specific working directory and \identifier{CODEX_HOME}, containing its configuration, Skill snapshot, and logs.
Native Skill loading requires filesystem reads; execution uses \identifier{danger-full-access} with interactive approval disabled.
Directories organize task state and artifacts without operating-system-level isolation.
Instructions permit read-only access to the prescribed Skill bundle but require scene perception and action through MCP tools, prohibiting filesystem or shell inspection of scene assets, ground-truth annotations, trajectories, and evaluator metadata.
The MCP whitelist restricts exposed environment tools; filesystem restrictions are instruction-level constraints.

\paragraph{Tool interfaces.}
The shared visual-point contract comprises a view, a normalized image point, and returned observations with execution feedback; concrete tool names and auxiliary operations depend on the backend.
Both variants expose \identifier{hab_init_scene}, \identifier{hab_panorama}, \identifier{hab_turn}, and \identifier{hab_close_session} for initialization, observation, in-place turning, and closure.
Learned Executor translates through \identifier{hab_local_navigate}, with \identifier{hab_localnav_status} and \identifier{hab_localnav_stop} for monitoring and cancellation.
Geo-based Executor uses \identifier{hab_visual_point_navigate}; valid requests execute without a preview--confirmation round trip.
The learned navigation call waits for the bounded hop to finish; status queries inspect a running or recent hop, and cancellation requests take effect at the next replanning boundary. These auxiliary operations support monitoring or interruption rather than being required after each completed call.
The Agent selects targets visually, while Geo-based Executor uses privileged geometry internally.

\paragraph{Returned observations and context.}
Initialization and navigation return four directional RGB images.
The global-task JSON filter removes absolute pose, maps, evaluator outputs, and selected artifact fields.
Validation or service failures may return only an error; panorama-capture failures are explicit.

\subsection{Skill Configuration}

The Skill snapshot manifest records file inventories, file hashes, and a bundle hash, checked before installation or reuse.
Codex discovers the bundle in a run-local Skills directory; the task prompt references the Skill without embedding its text, and generated instructions require reading its entry file before scene actions.
A visibility probe verifies the discovered name, description, and source path, blocking startup on failure; results may be reused for the same backend, entry Skill, and snapshot hash.
The entry files for Section~\ref{sec:skills} are listed below.

\begin{table}[!t]
\centering
\setlength{\tabcolsep}{2pt}
\begin{tabular}{@{}p{0.20\linewidth}p{0.39\linewidth}p{0.38\linewidth}@{}}
\textbf{Variant} & \textbf{Entry Skill} & \textbf{Tool-usage Skill} \\
Geo-based Executor & \identifier{global-navigation-geo-based-executor} & \identifier{localnav-pointnav-geo-based-executor} \\
Learned Executor & \review{\identifier{global-navigation-nomad}} & \identifier{localnav-pointnav} \\
Multi-goal & \review{\identifier{multi-global-navigation-nomad}} & \identifier{localnav-pointnav}
\end{tabular}
\end{table}

\subsection{Execution Limits and Evaluation}

\paragraph{Timeouts and session closure.}
Benchmark configurations set timeouts of 3,600\,s per Codex execution, 180\,s for MCP startup, and 300\,s per tool call.
The harness accepts an \emph{achieved} or \emph{blocked} outcome from \identifier{hab_close_session} only after successful closure.
It does not automatically resume a process lacking a valid terminal outcome: missing or unsuccessful closure is recorded separately, and the batch runner handles exceptions, including timeouts, before continuing to the next task.

\paragraph{Evaluation criteria.}
The Agent judges semantic completion using Skill guidance.
HM3D results use geodesic-distance criteria at the thresholds reported in Table~\ref{tab:hm3d_results}.
Our single-object and multi-object tasks use geodesic distance strictly below 1\,m to approved target viewpoints; demand-driven tasks retain 2\,m horizontal bounding-box regions.
All three require explicit STOP, with sequential tasks evaluated by ordered arrival along the trajectory.
Demand-driven scoring matches admissible candidate objects at each stage, without reusing an already matched candidate.
Appendix~\ref{sec:exp_details} defines the benchmark-specific scoring.


%% file: sections/experiment_details.tex
\section{Evaluation Protocols and Additional Results}
\label{sec:exp_details}

\begin{reviewblock}
\subsection{Real-World Navigation Tasks and Outcomes}
\label{sec:realworld_task_results}

SuperNav, LightNav-0, and ApexNav all use the same Unitree Go2 Air robot platform.
SuperNav and LightNav-0 use GoPro HERO8 Black cameras, while ApexNav uses an Intel RealSense D435i camera.

Table~\ref{tab:realworld_task_results} reports the ten real-world task instructions and the outcomes for SuperNav, ApexNav, and LightNav-0.
The set contains seven single-target tasks and three ordered two-target tasks.
SuperNav succeeds on nine tasks; its unsuccessful task is finding the elevator followed by the water dispenser.
Figures~\ref{fig:realworld_tasks_1_2}--\ref{fig:realworld_tasks_9_10} show the recorded trajectories for all ten tasks, ordered by the task IDs in Table~\ref{tab:realworld_task_results}.

\begin{table}[!t]
\color{black}
\centering
\caption{\textbf{Per-task real-world navigation outcomes.} S: success; F: failure.}
\label{tab:realworld_task_results}
\small
\setlength{\tabcolsep}{4pt}
\begin{tabular}{@{}rp{0.48\linewidth}ccc@{}}
\toprule
ID & Task instruction & SuperNav & ApexNav & LightNav-0 \\
\midrule
1 & Find the elevator. & S & F & F \\
2 & Find the basketball. & S & S & F \\
3 & Find another Go2 dog & S & F & F \\
4 & Find the printer, then the trash bin. & S & F & F \\
5 & Find the basketball, then the water bottle. & S & F & F \\
6 & Find the elevator, then the water dispenser. & F & F & F \\
7 & Find the chair. & S & S & S \\
8 & Find the trash bin. & S & S & S \\
9 & Find the printer. & S & F & S \\
10 & Find the broom. & S & F & F \\
\midrule
& Successful tasks & 9/10 & 3/10 & 3/10 \\
\bottomrule
\end{tabular}
\end{table}
\end{reviewblock}

\subsection{Instance-Navigation Benchmark Construction}
\label{sec:gs_construction}
We build our instance-navigation benchmark in the Habitat-GS simulator~\review{\citep{xia2026habitat}} using InteriorGS scene assets~\review{\citeinteriorgs}, generate object-level semantic annotations with a vision-language model, and manually review the annotations.
The final evaluation set contains 150 single-object tasks and 150 ordered multi-object tasks; the latter contain two to five targets per instruction.

\subsection{Instance- and Demand-Driven Navigation Metrics}
\label{sec:local_metrics}

\paragraph{Instance-navigation arrival and termination.}
For target $g$, let $V_g$ be its approved viewpoint set and $d_{\mathrm{nav}}$ the shortest geodesic distance on the scene's navigation mesh.
Arrival at position $p$ requires
\[
\min_{v\in V_g}d_{\mathrm{nav}}(p,v)<1\,\mathrm{m}.
\]
Single-object success requires the final saved position to satisfy this test and an explicit STOP.
Multi-object success requires saved trajectory positions to satisfy the targets in order, followed by explicit STOP; earlier targets need not remain near the final position.
We test the complete approved viewpoint set at each stage using Habitat's \identifier{MultiGoalShortestPath}, without interpolating between saved positions.
The initial position can satisfy a goal, and one position may satisfy consecutive stages.
Scoring is performed after execution and is not returned as policy feedback.

\paragraph{Viewpoint-reference SPL.}
For each task, SPL is $S_iL_i/\max(L_i,P_i)$, where $S_i$ is success under the 1\,m viewpoint criterion and $P_i$ is the full executed path length.
We use the dataset-provided \emph{global} reference: the shortest route over approved viewpoints in the fixed target order, computed by dynamic programming for multi-object tasks.
For single-object tasks, the local and global references coincide.
The reference reaches the viewpoint positions themselves; it is not shortened to the 1\,m arrival regions, and the denominator caps SPL at one when $P_i<L_i$.
All methods share the same reference for each task, and both SR and SPL average over all 150 tasks in each task set, each with a valid reference.
Failed tasks receive zero SPL; a successful zero-length reference and trajectory receive one.

\paragraph{Demand-driven navigation.}
We use tasks from Demand-Bench, introduced with DemandAgent~\citep{anonymous2026demandagent}, under an adapted evaluation protocol. DemandAgent evaluates explicit completion claims against functional dependencies and uses a 1.5\,m object-center or support-footprint criterion with applicable facing and visibility checks. Our evaluation instead scores ordered trajectory arrivals at admissible objects; its SR and SPL are therefore not directly comparable to DemandAgent\textquotesingle s reported scores.
Our protocol uses a horizontal bounding-box criterion: the distance from the robot to an admissible object's axis-aligned box must be at most 2\,m, including the box interior, without additional visibility, line-of-sight, or vertical-distance tests.
Success requires ordered completion of all stages and explicit STOP.
At each stage, the scorer selects the nearest unused admissible object, breaking ties by object ID; one position may complete consecutive stages, including at initialization.
SPL uses reference paths through all ordered 2\,m success regions, averaged over all 200 tasks, with the same references for every method.

\paragraph{Stage denominators and task composition.}
SR$_k$ counts trajectories completing the first $k$ goals among tasks with at least $k$ goals and does not require STOP.
Multi-object tasks contain two, three, four, or five goals in groups of 38, 37, 37, and 38 tasks, respectively.
Demand-driven tasks contain one through eight goals in groups of 70, 48, 45, 26, 8, 1, 1, and 1 tasks.
No method completes six or more stages; only three, two, and one tasks are eligible for stages six, seven, and eight, respectively.

\subsection{Ordered Goal Completion}
\label{sec:exp_progress}

Table~\ref{tab:stage_success} reports ordered progress using viewpoint geodesic distance below 1\,m for multi-object tasks and 2\,m bounding-box regions for demand-driven tasks.
SuperNav leads every multi-object stage and the first five demand-driven stages.

\begin{table}[!t]
\centering
\caption{\textbf{Ordered progress on multi-object and demand-driven tasks.} Multi-object tasks use viewpoint geodesic distance $<1$\,m; demand-driven tasks use 2\,m bounding-box regions. SR$_k$ measures completion of the first $k$ goals among tasks with at least $k$ goals, without requiring STOP. Overall task SR appears in Table~\ref{tab:local_navigation}.}
\label{tab:stage_success}
\small
\setlength{\tabcolsep}{3.5pt}
\begin{tabular}{@{}lrrrrrrrr@{}}
\toprule
\multicolumn{9}{l}{\textit{Multi-object navigation}} \\
Method & SR$_1\uparrow$ & SR$_2\uparrow$ & SR$_3\uparrow$ & SR$_4\uparrow$ & SR$_5\uparrow$  & & & \\
Eligible tasks & 150 & 150 & 112 & 75 & 38  & & & \\
\midrule
NaVid & 40.00 & 6.67 & 2.68 & 1.33 & 0.00 & & & \\
UniNaVid & 36.67 & 4.00 & 1.79 & 1.33 & 0.00 & & & \\
StreamVLN & 22.00 & 3.33 & 0.89 & 0.00 & 0.00 & & & \\
OmniNav (Action Former) & 44.00 & 11.33 & 1.79 & 1.33 & 0.00 & & & \\
SuperNav + Learned Executor & \textbf{86.00} & \textbf{63.33} & \textbf{52.68} & \textbf{42.67} & 34.21 & & & \\
SuperNav + Geo-based Executor & 85.33 & 62.00 & 48.21 & 38.67 & \textbf{39.47} & & & \\
\midrule
\multicolumn{9}{l}{\textit{Demand-driven navigation}} \\
Method & SR$_1\uparrow$ & SR$_2\uparrow$ & SR$_3\uparrow$ & SR$_4\uparrow$ & SR$_5\uparrow$ & SR$_6\uparrow$ & SR$_7\uparrow$ & SR$_8\uparrow$ \\
Eligible tasks & 200 & 130 & 82 & 37 & 11 & 3 & 2 & 1 \\
\midrule
NaVid & 28.50 & 11.54 & 7.32 & 0.00 & 0.00  & 0.00 & 0.00 & 0.00 \\
UniNaVid & 40.00 & 20.77 & 14.63 & 2.70 & 0.00  & 0.00 & 0.00 & 0.00 \\
StreamVLN & 45.00 & 23.08 & 12.20 & 2.70 & 0.00  & 0.00 & 0.00 & 0.00 \\
OmniNav (Action Former) & 48.50 & 26.92 & 15.85 & 8.11 & 0.00  & 0.00 & 0.00 & 0.00 \\
SuperNav + Learned Executor & 55.00 & 33.08 & 26.83 & 5.41 & 0.00  & 0.00 & 0.00 & 0.00 \\
SuperNav + Geo-based Executor & \textbf{72.00} & \textbf{55.38} & \textbf{45.12} & \textbf{24.32} & \textbf{9.09}  & 0.00 & 0.00 & 0.00 \\
\bottomrule
\end{tabular}
\end{table}

\subsection{Observation Interfaces and Baseline Execution}
\label{sec:baseline_inputs}

\begin{table}[!t]
\centering
\caption{\textbf{Baseline observation settings in Habitat-GS and AI2-THOR.} Dimensions refer to environment-rendered images before each model's preprocessing. Camera height is 1.25\,m. All policies receive the complete instruction and retain their native history mechanisms.}
\label{tab:baseline_inputs}
\small
\begin{tabular}{@{}lllll@{}}
\toprule
Method & RGB views & Resolution per view & HFOV & Precision \\
\midrule
NaVid & Front & $640\times480$ & $90^\circ$ & FP16 \\
UniNaVid & Front & $640\times480$ & $120^\circ$ & FP16 \\
StreamVLN & Front & $640\times480$ & $79^\circ$ & BF16 \\
OmniNav (Action Former) & Left, front, right & $720\times640$ & $110^\circ$ & BF16 \\
\bottomrule
\end{tabular}
\end{table}

\begingroup\raggedright
The baseline configurations preserve model-specific inputs rather than imposing a shared camera interface (Table~\ref{tab:baseline_inputs}).
NaVid uses the \identifier{navid-7b-full-224-video-fps-1-grid-2-r2r-rxr-training-split} checkpoint, and UniNaVid uses \identifier{uninavid-7b-full-224-video-fps-1-grid-2}.
StreamVLN uses its local checkpoint and native adapter.
OmniNav uses the \identifier{chongchongjj/OmniNav} fast-system checkpoint with its waypoint and arrival-prediction interface, including the required self-pose input.
Its positive arrival prediction is converted to STOP; for the video-policy adapters, a parsing-failure fallback is not accepted as an explicit STOP.
No Agentic controller, goal-wise policy reset, or evaluator stage-progress signal is added to these baselines.
\par\endgroup

SuperNav uses direction-labeled four-view observations.
In InteriorGS evaluation, the MLLM receives images at $256\times192$ resolution.
For single-object navigation, Geo-based Executor uses depth and navigation-mesh geometry; for demand-driven navigation, it uses selected-pixel depth and a reachable-position graph.
Neither grants the upper-level Agent access to evaluator targets.
The motion-backend comparison changes both the controller and its supporting geometric resources.

\subsection{Published HM3D Protocols}
\label{sec:hm3d_protocol}

\paragraph{HM3D-OVON.}
The public MTU3D and AstraNav-Memory navigation loops snap predicted destinations to the simulator navigation mesh and construct a \identifier{GreedyGeodesicFollower} with \identifier{PathFinder} to generate executable actions.
OmniNav's slow system uses the same mechanism in its \identifier{A-star} branch.
These are uses of simulator geometry for motion execution, independently of its use for computing evaluation distances.
We therefore use Geo-based Executor as our primary OVON comparison configuration.
Our OVON evaluation uses a 120-episode val-unseen subset.
The MTU3D, AstraNav-Memory, and OmniNav slow-system implementations compute geodesic distance to the nearest goal viewpoint, using thresholds of 0.25\,m, 1\,m, and 1\,m, respectively.
Their evaluation scripts continue the outer navigation loop while cumulative action counts are below 500, 2,000, and 4,000, respectively.
SoftNav reports a 0.25\,m threshold and 120 episodes per split.

\paragraph{HM3Dv2.}
The ApexNav table reports 1,000 validation episodes across 36 scenes and six target categories, with a 500-step limit and a 0.2\,m threshold.
L3MVN, VLFM, and SG-Nav values are taken from its HM3Dv2 reevaluations.
ApexNav's Habitat success measure requires STOP and geodesic distance strictly below 0.2\,m to a goal viewpoint.
WMNav~\citep{wmnav} and MSGNav~\citep{msgnav} report results on 1,000 HM3Dv2 validation episodes; we use WMNav's full-model result from Table II and MSGNav's arXiv v5 result from Table 2.
Their implementations test geodesic distance strictly below 1\,m to the nearest goal viewpoint, but do not require a Habitat STOP flag for success: WMNav scores after STOP or budget exhaustion, and MSGNav scores after its navigation loop exits.
WMNav's default 40-step budget counts its action loop rather than Habitat atomic actions, so it is not directly equivalent to a 500-action budget.

\subsection{OVON Evaluation and Ablation Implementation}
\label{sec:ovon_ablation_details}

\paragraph{Shared episodes and scoring.}
The four Skills/Tool-design configurations and the Learned Executor variant use the same 120-episode OVON val-unseen manifest, GPT-5.6 Terra in Codex user mode with high reasoning effort, and a 3,600\,s budget per episode; every episode ends with its session closed.
Success requires the final position's shortest navmesh geodesic distance to the goal-viewpoint set to be strictly below 0.25\,m or 1\,m, without a separate Habitat STOP requirement or reliance on the Agent's verbal judgment.
SPL averages $S_iL_i/\max(P_i,L_i)$ over all episodes, using the start-to-viewpoint shortest geodesic distance $L_i$ and executed path length $P_i$.
Following MTU3D's designated unreachable-goal rule, two episodes per configuration receive SR and SPL contributions of one.

\begin{reviewblock}
\paragraph{HarnessVLN comparison.}
We evaluate HarnessVLN~\citep{harnessvln} on the same 120-episode OVON val-unseen subset with the same evaluation settings as SuperNav, including a 3,600\,s timeout per episode.
At the 1\,m success threshold, HarnessVLN succeeds on 64 of 120 episodes, yielding 53.33\% SR and 0.2941 SPL (Table~\ref{tab:hm3d_results}).
\end{reviewblock}

\paragraph{Decision-model substitution.}
The model variant replaces GPT-5.6 Terra with high reasoning effort by GPT-6 Astra with medium reasoning effort.
It retains the full system's Navigation Skills, agent-oriented Tools, direct visual-point interface, Geo-based Executor, episode manifest, time budget, and scoring protocol.
Astra succeeds on 86 of 120 episodes at 0.25\,m and 95 of 120 at 1\,m.
This comparison varies the model and its stated reasoning setting while holding the navigation harness and execution configuration fixed.

\paragraph{Removing Navigation Skills.}
We disable global and local navigation Skills but retain system Skills and MCP interface descriptions and schemas.
The prompt contains only the goal and an instruction to close the session; generated agent guidance is empty, and repository-document injection and shell access are disabled.
The descriptions specify initialization order, normalized image coordinates, feedback, and closure, without search, recovery, or final-approach strategies.
The runner preconfigures the scene and starting pose, allowing argument-free initialization.

\paragraph{Ablating agent-oriented Tool design.}
The four Skills/Tool-design configurations expose the same initialization, turning, visual-point navigation, and closing tools, retaining depth back-projection, reachable-navmesh snapping, and follower execution.
The full interface returns front/right/back/left RGB after initialization, movement, and turning, and accepts points on any latest directional image.
The front-only interface returns one forward image, removes view-selection and image-reference parameters, and binds points to the latest forward image, rejecting other-direction or stale references.
The Agent must turn before inspecting or pointing sideways or backward; no panorama tool or scanning schedule is provided.

\paragraph{Adapting Skills.}
Front-only Skills retain target judgment, exploration-branch records, exploration and return, recovery, final approach, and termination procedures.
They replace four-view inspection with forward-view inspection and require turning before side or rear pointing; the Agent chooses when and where to look.
Both Skill versions require final pointing at the target or its base followed by close-range visual verification, with approximately 1\,m distance guidance.
The Learned Executor variant uses its own global and local Skills and four-view RGB; its results appear only in the backend comparison (Table~\ref{tab:hm3d_results}).

\paragraph{Language grounding versus direct point selection.}
\label{par:grounding_motion_tool}
Grounding Motion Tool denotes our interface built on LocateAnything~\review{\citep{locateanything}} (\identifier{visual_ground_preview}).
The Agent specifies a visible object or proxy with a natural-language referring expression in a selected camera view.
An external grounding model localizes the referent, and the tool uses depth and navigability checks to construct candidate navigation goals.
A unique candidate that requires no confirmation is executed directly; multiple candidates and designated doorway candidates require the Agent to select and confirm a candidate from a numbered visual overlay.
In contrast, direct point navigation requires the Agent to predict a normalized image coordinate itself; the tool projects this pixel using depth and executes navigation toward a reachable location without a preview--confirmation step.
Both variants use the same underlying navigation controller, while differing in how visual navigation goals are specified and selected.
This comparison examines external language-conditioned grounding and candidate selection relative to the Agent's own visual point selection.

\begin{table}[!t]
\centering
\caption{\textbf{Interaction and execution statistics for Geo-based Executor ablations.} Means include all 120 episodes, both successes and failures. Calls count Habitat tools, including unsuccessful calls; turns may serve observation or heading adjustment.}
\label{tab:ovon_ablation_execution}
\small
\setlength{\tabcolsep}{4pt}
\begin{tabular}{@{}lrrrr@{}}
\toprule
Variant & Calls & Navigation & Turns & Path (m) \\
\midrule
w/o Navigation Skills & 7.43 & 5.33 & 0.10 & 7.56 \\
w/o Agent-oriented Tool Design & 30.28 & 10.72 & 17.56 & 23.60 \\
w/o Skills and Tool Design & 15.37 & 5.61 & 7.72 & 10.47 \\
\midrule
SuperNav (Full) & 11.17 & 8.76 & 0.41 & 16.44 \\
\bottomrule
\end{tabular}
\end{table}

Tool-call counts exclude shell operations for reading Skills and non-Habitat resource-listing calls.
In the no-Skills, front-only configuration, one episode initializes five sessions; its score and path length use the final session associated with the reported metrics, whereas tool-call counts include the entire run.

\begin{reviewblock}
\subsection{InstructNav-adapted: Fixed-Sequence Navigation Baseline}
\label{sec:instructnav_adapted}

\paragraph{Shared configuration.}
We adapt InstructNav~\citep{instructnav} to the same 120-episode OVON val-unseen subset, GPT-5.6 Terra model with high reasoning effort, four-camera observation interface, and Geo-based Executor motion executor used by SuperNav.
Each episode has a 3,600\,s budget covering perception, model requests, retries, and motion.
Both GPT calls in each iteration use \identifier{gpt-5.6-terra}, high reasoning effort, and \identifier{summary=auto}; their order is fixed by the program.

\paragraph{Decision loop.}
Each iteration executes the following sequence:
\begin{enumerate}
\item Acquire front, right, back, and left RGB-D observations and poses through SuperNav's observation interface.
\item Apply GLEE~\citep{wu2023GLEE} with its official weights, project semantic masks into 3D point clouds using depth, and update maps of objects, traversable regions, obstacles, and exploration frontiers.
\item Query GPT with the task, plan history, and observed categories to generate a DCoN \texttt{Action--Landmark--Flag} tuple.
\item Query GPT with the four RGB images, task, and current DCoN to select a viewing direction or issue \texttt{Stop}.
\item Compute semantic, action, trajectory, and intuition value maps, each weighted by 0.25 during ordinary operation. Apply obstacle constraints, select the highest-value location, and compute an A* path.
\item Convert a visible location along the path into image coordinates and execute motion through SuperNav's \identifier{visual_point} interface. After the shared motion call completes, restart the decision loop.
\end{enumerate}

\paragraph{Adaptation details.}
The visual reasoning interface uses four directions in place of the official implementation's default six directions.
Camera calibration follows SuperNav, with the ground threshold adjusted to the actual camera height.
Prompts are rewritten for four-view input, JSON output, and the current task while preserving the roles of DCoN generation and visual-direction reasoning.
For execution, A* path locations are converted into visible image markers and passed to the shared motion planner.
Replanning occurs after every shared motion call; the official implementation primarily triggers replanning based on traveled distance and follower state.

\paragraph{Termination and evaluation.}
An episode terminates when GPT outputs \texttt{Stop}, when a valid \texttt{Flag} is present and the distance to the map goal is below 0.5\,m, or when the time budget expires.
If no valid visible path exists, the system performs one frontier replanning attempt; continued failure terminates the episode with \texttt{blocked}.
The internal 0.5\,m stopping condition is separate from the reported evaluation thresholds of strictly below 0.25\,m and 1\,m.
The method succeeds on 50/120 and 65/120 episodes at these thresholds, yielding SR of 41.67\% and 54.17\% and SPL of 0.1865 and 0.2306, respectively (Table~\ref{tab:hm3d_results}).
We label this configuration InstructNav-adapted to identify its changes to the motion interface, replanning frequency, and termination rules.

\subsection{Execution Time and Token Usage}
\label{sec:runtime_tokens}

\paragraph{Model access and latency.}
SuperNav's HM3Dv2 and custom-benchmark evaluations used model requests routed through a third-party API relay, with higher request latency; the OVON ablations and Astra evaluations used an official Codex subscription, with lower model latency.
Cache-hit behavior may also differ between these access routes.
The reported execution times therefore reflect both navigation behavior and the model-serving conditions of each evaluation.

\paragraph{Execution time.}
Table~\ref{tab:benchmark_runtime} reports mean episode execution times on the instance-navigation and demand-driven benchmarks.
Table~\ref{tab:hm3d_runtime} reports episode-time distributions, cumulative model time, and tool-call statistics on HM3Dv2 and OVON.
All durations are in seconds; tool-call rates are mean calls divided by mean episode execution time.
Terra denotes GPT-5.6 Terra (high), and Astra denotes GPT-6 Astra (medium).
The OVON Skills/Tool-design ablations use Geo-based Executor.

\begin{table}[!t]
\color{black}
\centering
\caption{\textbf{Mean episode execution time on instance-navigation and demand-driven benchmarks.} All values are in seconds. Task counts are in parentheses.}
\label{tab:benchmark_runtime}
\small
\setlength{\tabcolsep}{5pt}
\begin{tabular}{@{}lrrr@{}}
\toprule
Method & Single-object (150) & Multi-object (150) & Demand-driven (200) \\
\midrule
NaVid & 266.40 & 240.60 & 98.40 \\
UniNaVid & 67.80 & 40.80 & 56.40 \\
StreamVLN & 30.00 & 23.40 & 45.60 \\
OmniNav (Action Former) & 112.80 & 153.00 & 258.60 \\
\midrule
SuperNav + Learned Executor & 577.80 & 2610.60 & 1786.80 \\
SuperNav + Geo-based Executor & 384.00 & 2472.60 & 539.40 \\
\bottomrule
\end{tabular}
\end{table}

\begin{table}[!t]
\color{black}
\centering
\caption{\textbf{Execution time and tool usage on HM3Dv2 and OVON.} Mean, median, and P95 describe episode execution time in seconds. Model time is the mean cumulative model time per episode in seconds; calls are mean tool calls per episode.}
\label{tab:hm3d_runtime}
\scriptsize
\setlength{\tabcolsep}{3pt}
\begin{tabular}{@{}lrrrrrrr@{}}
\toprule
Configuration & Episodes & Mean (s) & Median (s) & P95 (s) & Model (s) & Calls & Calls/s \\
\midrule
\multicolumn{8}{l}{\textit{HM3Dv2}} \\
Geo-based Executor & 1,000 & 445.25 & 274.88 & 1410.20 & 383.05 & 15.20 & 0.0341 \\
Learned Executor & 1,000 & 281.67 & 199.28 & 751.10 & 185.36 & 17.68 & 0.0628 \\
\midrule
\multicolumn{8}{l}{\textit{OVON}} \\
Geo-based Executor (Full) & 120 & 174.78 & 125.57 & 459.42 & 103.95 & 11.17 & 0.0639 \\
w/o Navigation Skills & 120 & 66.94 & 53.37 & 153.30 & 33.18 & 7.43 & 0.1110 \\
w/o Agent-oriented Tool Design & 120 & 285.77 & 208.71 & 890.81 & 168.32 & 30.28 & 0.1060 \\
w/o Skills and Tool Design & 120 & 106.69 & 66.67 & 292.39 & 57.54 & 15.37 & 0.1441 \\
Learned Executor (Full) & 120 & 315.06 & 265.25 & 757.19 & 171.38 & 18.19 & 0.0577 \\
\midrule
\multicolumn{8}{l}{\textit{OVON, Astra}} \\
Geo-based Executor (Full) & 120 & 107.85 & 85.62 & 220.09 & 52.97 & 8.87 & 0.0822 \\
\bottomrule
\end{tabular}
\end{table}

\paragraph{Token usage.}
Table~\ref{tab:hm3d_tokens} reports mean recorded token usage per episode in thousands of tokens (k tokens) across the navigation benchmarks.
Cached input is included in total input, and reasoning tokens are included in total output.

\begin{table}[!t]
\color{black}
\centering
\caption{\textbf{Mean recorded token usage per episode across navigation benchmarks.} All token counts are in k tokens. Cached input and reasoning are subsets of input and output, respectively.}
\label{tab:hm3d_tokens}
\small
\setlength{\tabcolsep}{4pt}
\begin{tabular}{@{}lrrrr@{}}
\toprule
Configuration & Input & Cached input & Output & Reasoning \\
\midrule
\multicolumn{5}{l}{\textit{HM3Dv2}} \\
Geo-based Executor & 1058.40 & 984.07 & 5.99 & 3.34 \\
Learned Executor & 1038.02 & 976.24 & 7.60 & 4.53 \\
\midrule
\multicolumn{5}{l}{\textit{OVON}} \\
Geo-based Executor (Full) & 588.99 & 543.28 & 4.99 & 2.70 \\
w/o Navigation Skills & 163.45 & 140.51 & 1.40 & 0.60 \\
w/o Agent-oriented Tool Design & 1189.87 & 1128.02 & 6.92 & 3.86 \\
w/o Skills and Tool Design & 345.99 & 314.93 & 2.19 & 0.88 \\
Learned Executor (Full) & 950.14 & 894.92 & 8.14 & 4.83 \\
\midrule
\multicolumn{5}{l}{\textit{OVON, Astra}} \\
Geo-based Executor (Full) & 358.76 & 330.90 & 1.93 & 0.22 \\
\midrule
\multicolumn{5}{l}{\textit{Single-object navigation}} \\
Geo-based Executor & 929.07 & 866.91 & 5.71 & 3.14 \\
Learned Executor & 858.56 & 793.79 & 7.80 & 4.71 \\
\midrule
\multicolumn{5}{l}{\textit{Multi-object navigation}} \\
Geo-based Executor & 2013.81 & 1911.35 & 10.25 & 6.16 \\
Learned Executor & 2272.45 & 2167.10 & 12.95 & 8.21 \\
\midrule
\multicolumn{5}{l}{\textit{Demand-driven navigation}} \\
Geo-based Executor & 202.33 & 168.81 & 1.83 & 0.97 \\
Learned Executor & 434.17 & 384.37 & 2.97 & 1.73 \\
\bottomrule
\end{tabular}
\end{table}
\end{reviewblock}

\begin{reviewblock}
\subsection{Motion Executors and Path-Efficiency Analysis}
\label{sec:motion_path_efficiency}

\paragraph{Map and planner audit.}
Table~\ref{tab:motion_executor_audit} summarizes the map sources and execution mechanisms in the public simulation implementations of the methods in Table~\ref{tab:hm3d_results}.
GT depth denotes simulator-rendered depth, and GT pose denotes simulator GPS/compass, camera poses, or pose increments derived from simulator states.
These observations differ from access to the complete GT NavMesh for path search.

\begin{table}[!t]
\color{black}
\centering
\caption{\textbf{Map sources and path-planning mechanisms.} The audit covers public simulation implementations and default launchers. SoftNav's executor remains unverified because attributable public execution code was not located.}
\label{tab:motion_executor_audit}
\footnotesize
\setlength{\tabcolsep}{5pt}
\renewcommand{\arraystretch}{1.15}
\begin{tabular}{@{}>{\raggedright\arraybackslash}p{\dimexpr0.24\linewidth-2\tabcolsep\relax}>{\raggedright\arraybackslash}p{\dimexpr0.35\linewidth-2\tabcolsep\relax}>{\raggedright\arraybackslash}p{0.41\linewidth}@{}}
\toprule
\textbf{Method} & \textbf{Map or geometric information} & \textbf{Planning and execution} \\
\midrule
MTU3D\newline\citep{mtu3d} & GT NavMesh; simulator top-down map for exploration & Goal snapping; GreedyGeodesicFollower (GGF) and atomic actions \\
\addlinespace[3pt]
SoftNav\newline\citep{softnav} & Unverified & Unverified \\
\addlinespace[3pt]
AstraNav-Memory\newline\citep{astranav} & GT NavMesh & Goal snapping; GGF and atomic actions \\
\addlinespace[3pt]
OmniNav (slow + CoT)\newline\citep{omninav} & GT NavMesh in the default slow-system launcher & The \identifier{A-star} branch calls goal snapping and GGF \\
\addlinespace[3pt]
L3MVN\newline\citep{l3mvn} & Online semantic/occupancy maps from GT depth and pose & FMM short-term goals; deterministic atomic actions \\
\addlinespace[3pt]
VLFM\newline\citep{vlfm} & Online obstacle/frontier maps from GT depth and GPS/compass & Map-based frontier selection; learned PointNav uses GT depth and relative goal coordinates from GT pose \\
\addlinespace[3pt]
SG-Nav\newline\citep{sgnav} & Online semantic/free-space maps from GT depth and pose & FMM; deterministic atomic actions \\
\addlinespace[3pt]
ApexNav\newline\citep{apexnav} & Online occupancy map from GT depth and pose & A* on the online map; atomic actions \\
\addlinespace[3pt]
WMNav\newline\citep{wmnav} & Default depth/pose for candidate navigability and map updates & VLM-selected polar primitives; NavMesh \identifier{try_step} constrains local displacement, not global path search \\
\addlinespace[3pt]
MSGNav\newline\citep{msgnav} & Online TSDF/scene graph from GT depth and sensor poses; GT NavMesh for execution & \identifier{pathfinder.find_path}; waypoint teleportation through \identifier{agent.set_state} \\
\addlinespace[3pt]
HarnessVLN\newline\citep{harnessvln} & Online map from GT depth and pose increments & FMM by default; learned NavDP~\citep{cai2025navdp} under the stair condition \\
\addlinespace[3pt]
InstructNav-adapted\newline\citep{instructnav} & Online RGB-D semantic/traversability/frontier/value maps; GT NavMesh for execution & Online A* selects a visible location; shared SuperNav GT NavMesh/GGF executes it \\
\midrule
SuperNav\newline+ Geo-based Executor & Rendered depth, camera poses, and precomputed GT NavMesh & Depth back-projection and NavMesh snapping; \identifier{find_path} and GGF \\
\addlinespace[3pt]
SuperNav\newline+ Learned Executor & No planning map; marked reference image, current RGB, and visual history & Learned local trajectories converted into forward/turn primitives \\
\bottomrule
\end{tabular}
\end{table}

\paragraph{Executor comparison on OVON.}
We compare three motion executors within SuperNav on the same 120-episode OVON val-unseen subset: Geo-based Executor, the ApexNav executor~\citep{apexnav}, and Learned Executor.
The ApexNav configuration replaces only the motion executor, retaining the same high-level agent and evaluation settings.
Table~\ref{tab:ovon_executor_comparison} reports SR and SPL over all 120 episodes.
Table~\ref{tab:ovon_executor_joint_success} reports mean SPL over the episodes successful under all three executors at each threshold.

\begin{table}[!t]
\color{black}
\centering
\caption{\textbf{Motion executor comparison within SuperNav on OVON.} All configurations use the same 120-episode val-unseen subset. SR (\%) and SPL (0--1) use the indicated success thresholds; bold marks the highest value in each column.}
\label{tab:ovon_executor_comparison}
\begingroup
\small
\setlength{\tabcolsep}{6pt}
\begin{tabular}{@{}lrrrr@{}}
\toprule
& \multicolumn{2}{c}{0.25\,m} & \multicolumn{2}{c}{1\,m} \\
\cmidrule(lr){2-3}\cmidrule(l){4-5}
Motion executor & SR$\uparrow$ & SPL$\uparrow$ & SR$\uparrow$ & SPL$\uparrow$ \\
\midrule
Geo-based Executor & \textbf{68.33} & 0.3837 & \textbf{73.33} & 0.4105 \\
ApexNav Executor & 65.00 & \textbf{0.3927} & 70.83 & \textbf{0.4281} \\
Learned Executor & 65.00 & 0.1355 & 70.83 & 0.1443 \\
\bottomrule
\end{tabular}
\par\endgroup

\medskip
\caption{\textbf{SPL on OVON episodes successful under all three executors.} At each threshold, all columns are averaged over the same common success set. Higher is better; bold marks the highest value in each row.}
\label{tab:ovon_executor_joint_success}
\small
\setlength{\tabcolsep}{6pt}
\begin{tabular}{@{}lrrr@{}}
\toprule
Success threshold & ApexNav Executor & Geo-based Executor & Learned Executor \\
\midrule
0.25\,m & \textbf{0.6554} & 0.6041 & 0.2317 \\
1\,m & \textbf{0.6594} & 0.5884 & 0.2208 \\
\bottomrule
\end{tabular}
\end{table}

\paragraph{How execution affects SPL.}
For successful episodes with a fixed reference length, SPL decreases as executed path length increases.
Access to the complete GT NavMesh supports shortest-path queries between selected locations, while episode-level path efficiency also depends on waypoint selection, local stopping, and subsequent replanning.
Using Geo-based Executor therefore does not guarantee higher SPL.
In the controlled OVON comparison, replacing Geo-based Executor with the ApexNav executor raises SPL from 0.3837 to 0.3927 at 0.25\,m and from 0.4105 to 0.4281 at 1\,m, with SR decreasing from 68.33\% to 65.00\% and from 73.33\% to 70.83\%, respectively (Table~\ref{tab:ovon_executor_comparison}).
ApexNav also achieves the highest SPL on the common success sets at both thresholds, showing that its path-efficiency advantage persists when all three executors complete the same tasks (Table~\ref{tab:ovon_executor_joint_success}).
Learned Executor has lower SPL on these same successful episodes, indicating longer paths relative to the reference shortest distances.
High-level decisions also affect performance with a fixed executor: InstructNav-adapted shares our GT NavMesh/GGF executor and uses different target-selection, replanning, and termination logic.

\paragraph{Analysis on jointly successful HM3Dv2 episodes.}
We analyze 693 HM3Dv2 episodes successful under both Geo-based Executor and Learned Executor, separating path inefficiency from differences in which episodes succeed.
Table~\ref{tab:joint_success_efficiency} reports trajectory length, local detouring, and repeated approaches.
Success-region re-entry means entering the 0.2\,m success region, moving at least 1\,m away, and subsequently entering it again.
The segment detour ratio divides traveled distance by the NavMesh shortest distance between segment endpoints.

\begin{table}[!t]
\color{black}
\centering
\caption{\textbf{Path efficiency on 693 jointly successful HM3Dv2 episodes.} Re-entry and additional travel use the 0.2\,m success region. Lower is better for all four measurements.}
\label{tab:joint_success_efficiency}
\small
\setlength{\tabcolsep}{5pt}
\begin{tabular}{@{}lrr@{}}
\toprule
Measurement & Geo-based Executor & Learned Executor \\
\midrule
Mean trajectory length (m) & \textbf{22.25} & 50.67 \\
Segment detour ratio & \textbf{1.01} & 1.44 \\
Success-region re-entry (\%) & \textbf{18.8} & 39.7 \\
Mean travel after first success-region entry (m) & \textbf{7.06} & 26.75 \\
\bottomrule
\end{tabular}
\end{table}

Some revisits support legitimate verification or recovery, and repeated turns alone are not necessarily inefficient because they can acquire observations without increasing traveled distance.
Nevertheless, the larger detours, repeated approaches, and travel after initial arrival show that the Learned Executor configuration's lower SPL reflects genuinely inefficient local behavior as well as necessary exploration.
Local stopping and its coordination with task-level verification are therefore concrete targets for improvement.
\end{reviewblock}

%% file: sections/method_localnav.tex
\section{Learned Executor: Point-Goal Navigation in Image Space}
\label{sec:localnav}
\label{sec:localnav_impl}

\begin{figure}[!t]
    \centering
    \includegraphics[page=1,width=\linewidth]{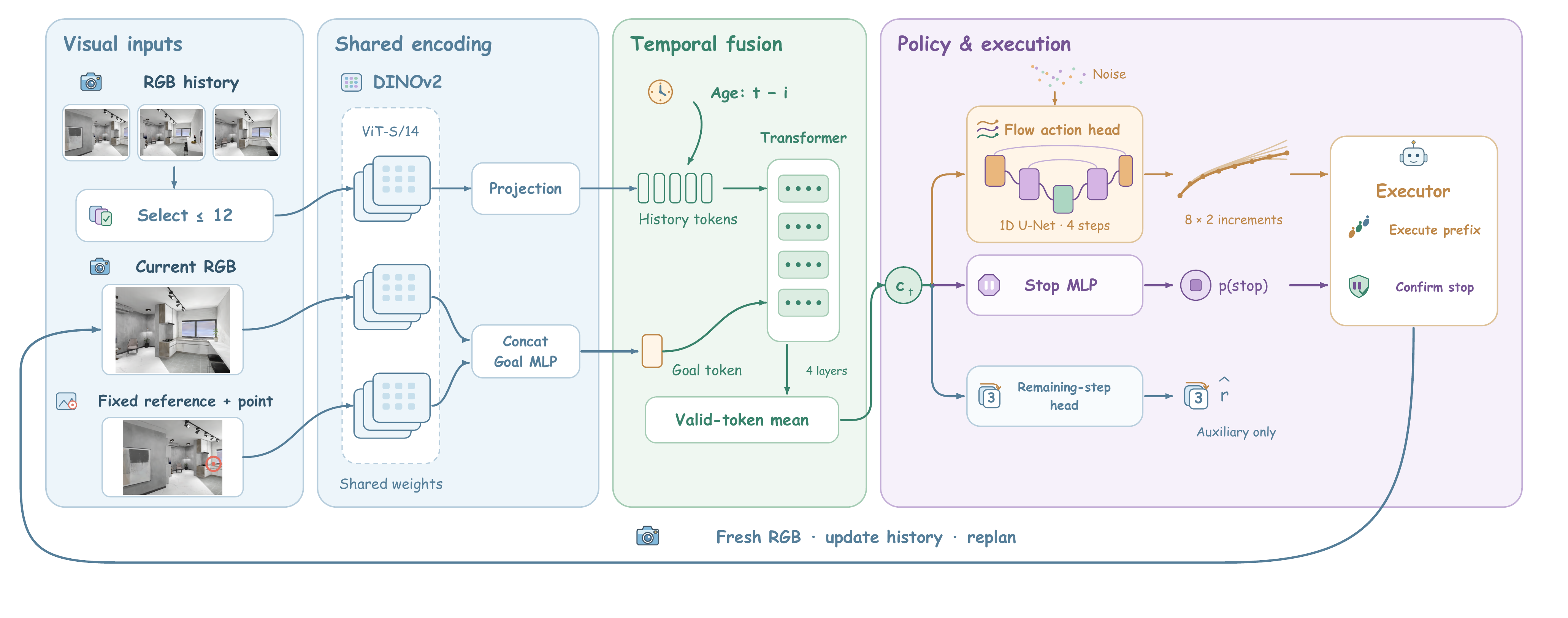}
    \caption{\textbf{Learned Executor architecture.} A point selected in a fixed reference image is fused with a bounded, age-aware RGB history. The policy predicts displacement increments and a stop score; the executor converts the increments to motion and returns fresh observations for replanning.}
    \label{fig:localnav_architecture}
\end{figure}

\paragraph{Overview.}
All experiments with a learned motion backend use the policy described in this appendix, including instance-level, multi-object, demand-driven, and HM3D navigation.
Learned Executor converts an image-space destination into a sequence of local movements as the robot's viewpoint changes.
The MLLM selects the destination and judges task completion, while Learned Executor predicts motion and when to stop the current movement request.
Inspired by NoMaD's goal-conditioned generative-policy design~\citep{nomad}, Learned Executor combines a marked reference image, a shared DINOv2 encoder, variable-length visual history, and a conditional flow head.
The marked image specifies a point within an observed scene, replacing the destination-image goal used by NoMaD; the flow head generates displacement increments conditioned on this point and the evolving visual history.
We train Learned Executor on simulator expert trajectories, initializing its visual encoder from pretrained DINOv2 and training the navigation policy independently of released NoMaD navigation weights.
Figure~\ref{fig:localnav_architecture} summarizes the resulting policy and execution loop.

\subsection{Visual Goal and History Encoding}

\paragraph{Point-goal representation.}
A local request specifies a reference RGB image $I^{\mathrm{ref}}$ and a normalized pixel $q=(u,v)\in[0,1]^2$ identifying the destination.
Both remain fixed throughout the request, while the current observation $I_t$ changes with motion.
RGB images are resized to $224\times224$; a red ring with a central dot is drawn at $q$ after resizing to obtain $\widetilde I^{\mathrm{ref}}=\mathrm{Mark}(I^{\mathrm{ref}},q)$.
Images are normalized using ImageNet statistics and encoded separately by a shared DINOv2 ViT-S/14 backbone $\phi$~\citep{dinov2}, producing 384-dimensional global features.
The observation projection $W_o$ and goal MLP produce 256-dimensional tokens:
\begin{equation}
    z_i=W_o\phi(I_i), \qquad
    g_t=\mathrm{MLP}_g\!\left([\phi(I_t);\phi(\widetilde I^{\mathrm{ref}})]\right).
    \label{eq:localnav_visual}
\end{equation}
The goal MLP has dimensions $768\rightarrow256\rightarrow256$, with a ReLU between its linear layers.
By combining the current and marked reference features, $g_t$ relates the fixed destination to the current view using RGB inputs.

\begin{figure}[!t]
    \centering
    \includegraphics[width=\linewidth]{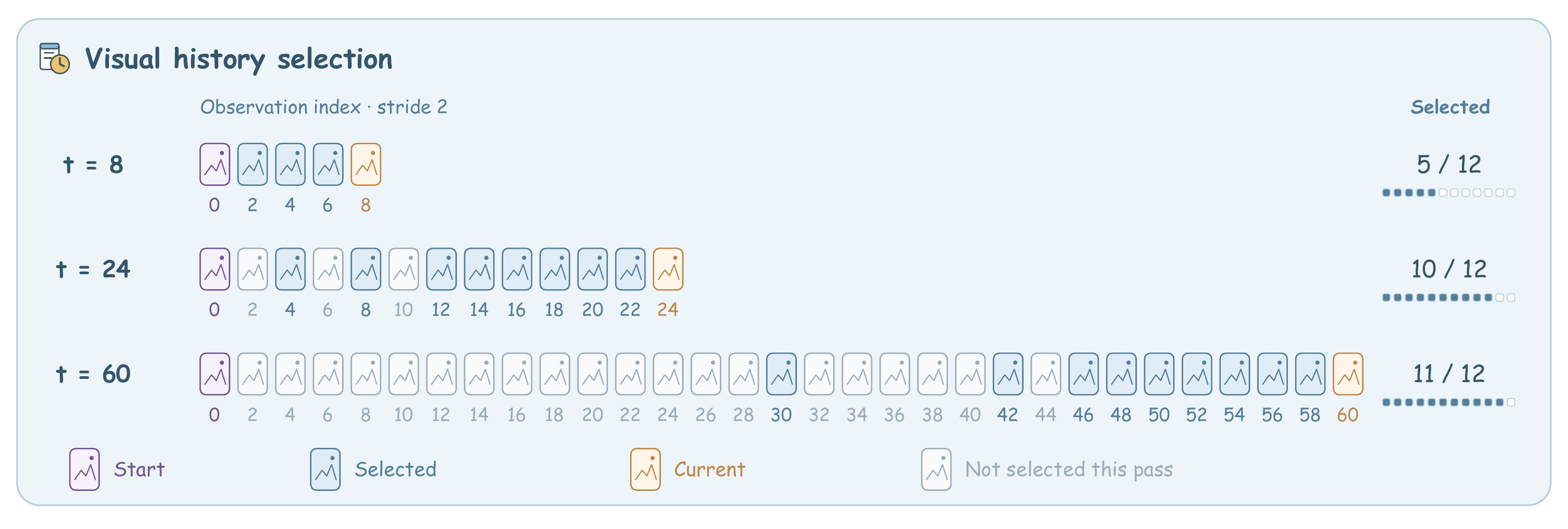}
    \caption{\textbf{Variable-length visual memory.} Each row shows every stride-two candidate on a shared observation-index scale, from the start to the current frame. Colored frames are selected; gray frames are omitted from this fusion pass but remain available in the history. Repeated thinning of the older candidate half limits the input to at most 12 observations, selecting 5, 10, and 11 frames at $t=8$, 24, and 60, respectively.}
    \label{fig:localnav_memory}
\end{figure}

\paragraph{Variable-length visual history.}
As the robot moves, the selected destination may leave its current view, making earlier observations useful context.
At observation index $t$, we select a history index set $\mathcal S_t$ containing the initial and current frames, with $K_t=|\mathcal S_t|\leq12$.
Selection starts from a stride-two sequence plus the current frame; when it exceeds the budget, we repeatedly thin the older half while preserving the endpoints.
This retains observations across the local request with finer sampling toward the present.
For example, the scheduler selects $[0,2,4,6,8]$ at $t=8$ and $[0,30,42,46,48,50,52,54,56,58,60]$ at $t=60$ (Figure~\ref{fig:localnav_memory}).
Because selected frames are not equally spaced, we encode each frame's age $t-i$ using a projected sinusoidal embedding $\psi$:
\begin{equation}
    h_i=z_i+\psi(t-i)+e_o, \qquad
    c_t=\mathrm{Mean}_{\mathrm{valid}}\!\left(
    \mathrm{Transformer}\big([\{h_i\}_{i\in\mathcal S_t};\,g_t+e_g]\big)\right).
    \label{eq:localnav_memory}
\end{equation}
Here $e_o$ and $e_g$ identify observation and goal tokens; ages use the recorded observation indices.
A four-layer, four-head Transformer with width 256 and feed-forward width 1,024 fuses the tokens.
Padding is excluded from attention and pooling, and the goal token participates in the mean.
At inference, projected image features are cached before age embeddings are added, so a selected frame can be reused with its updated age.
The 12-observation budget applies to each fusion pass, while the feature cache retains earlier observations within the local request.

\subsection{Motion Representation and Training}

\paragraph{Expert trajectories and point goals.}
We train Learned Executor by supervised imitation on Habitat greedy-follower trajectories, recording RGB observations and robot poses after motion primitives.
Future expert positions are projected into the initial image to construct visible point goals, with depth used to check occlusion during data generation.
For object-surface point samples, a nearby visible surface supplies the image marker while the corresponding ground-level endpoint supplies the approach trajectory.
The reference image and pixel remain fixed as each trajectory is expanded into training samples at successive current frames.
Training applies shared brightness augmentation to a sample's images and pixel jitter of up to two pixels per axis to its goal marker.

\paragraph{Displacement-increment representation.}
The policy predicts eight two-dimensional displacement increments, all expressed in the robot's coordinate frame at prediction time, with $x$ forward and $y$ left.
Let $P_j$ be the expert ground-plane position at frame $j$, $T$ the target-frame index, and $R_t$ the transformation from world-plane displacements to the robot frame at $t$.
The normalized training target $A\in\mathbb R^{8\times2}$ is defined by
\begin{equation}
    A_k=\mathrm{clip}\!\left(
    \frac{R_t\big(P_{\min(t+k,T)}-P_{\min(t+k-1,T)}\big)}{0.25\,\mathrm m},-1,1\right),
    \quad k=1,\ldots,8.
    \label{eq:localnav_action_target}
\end{equation}
Clipping is componentwise, and increments after the endpoint are zero.
Multiplying the predictions by 0.25\,m gives displacement increments in meters; summing these increments in the fixed prediction-time frame gives the corresponding relative waypoints.

\paragraph{Conditional flow prediction.}
A one-dimensional U-Net generates the increment sequence conditioned on the fused visual representation $c_t$.
Following flow matching~\citep{flowmatching}, we interpolate between the normalized sequence and Gaussian noise and regress the corresponding flow velocity:
\begin{equation}
    x_\tau=(1-\tau)A+\tau\epsilon, \qquad
    \mathcal L_{\mathrm{flow}}=
    \frac{1}{16}\,\mathbb E\!\left[\left\|v_\theta(x_\tau,\tau,c_t)-(\epsilon-A)\right\|_F^2\right],
    \label{eq:localnav_flow}
\end{equation}
where $\epsilon\sim\mathcal N(0,I)$ and $\tau\sim\mathcal U(0,1)$; the factor $1/16$ averages over the sequence components.
The flow network receives the rescaled time $10\tau$ internally.
For classifier-free guidance, 10\% of training samples replace the goal token with a learned null token while retaining the observation history.

\paragraph{Stopping supervision and optimization.}
Separate heads predict a stopping probability $p_t$ and an auxiliary remaining-step estimate $\hat r_t$ from $c_t$.
The supervision uses the expert observation-index gap $r_t=T-t$ and stopping label $y_t=\mathbf1[r_t\leq2]$; the index gap includes turning primitives.
We optimize $\mathcal L_{\mathrm{flow}}+0.5\mathcal L_{\mathrm{stop}}+0.002\mathcal L_{\mathrm{remaining}}$, using binary cross-entropy with positive-label weight 3 for stopping and squared error for the remaining index gap.
The flow loss averages over all samples and sequence components, while the stop and remaining-step losses average over samples retaining their goal token.
We train for 45,000 updates with global batch size 256 using AdamW, a learning rate of $10^{-4}$ for policy layers and $10^{-5}$ for the pretrained visual backbone, weight decay $10^{-6}$, and 500 warmup updates followed by cosine decay.
Training uses bfloat16 automatic mixed precision, gradient-norm clipping at 1, and an exponential moving average with decay 0.999.
The visual backbone is fine-tuned jointly with the policy, and inference uses the averaged weights.

\subsection{Closed-Loop Motion Execution}

\paragraph{Goal initialization and sequence sampling.}
For a new goal, the runtime saves the selected reference image and aligns the robot with that view's capture direction; alignment observations can also enter the history.
The current simulation configuration uses a camera height of 1.25\,m.
At each replanning call, sequence sampling starts with $x_1\sim\mathcal N(0,I)$ and uses four Euler steps,
$x_{\tau-1/4}=x_\tau-\frac14 v_{\mathrm{guided}}(x_\tau,\tau,c_t)$.
We use $v_{\mathrm{guided}}=v_u+2(v_c-v_u)$, where $v_c$ retains the goal token and $v_u$ uses the null goal with the same observation history.
The executor selects one of 16 sampled sequences using endpoint-direction agreement and path straightness.

\paragraph{Increment execution and replanning.}
The selected displacement increments are rescaled by 0.25\,m, and the executor normally follows a prefix of three increments.
The direction and length of each increment determine a heading change and travel distance, which the simulator adapter executes as turn-then-translate sequences using $10^{\circ}$ turns and 0.25\,m forward primitives.
A predicted increment can therefore require multiple motion primitives.
The executor then acquires new RGB observations and predicts another sequence toward the same reference goal.
Collision recovery and blockage detection run within the executor; in simulation, the blockage monitor uses recent robot positions to assess progress.

\paragraph{Local stopping and feedback.}
For this controller configuration, the executor confirms a local stop when $p_t\geq0.1$ on at least two consecutive replanning calls and the motion-progress condition is satisfied.
This condition requires either (i) at least four executed primitives in total and at least three since the first above-threshold prediction in the current streak, or (ii) no executed primitives in the previous cycle.
A below-threshold prediction resets the streak; the auxiliary remaining-step estimate does not gate stopping.
Execution continues until a confirmed stop, failure, or budget limit.
Failure and budget termination are reported separately from a confirmed local stop, which ends only the current movement request.
The Tool returns execution status and fresh views; the MLLM then assesses task completion and chooses the next operation.

%% file: sections/realworld_trajectories.tex
\clearpage
\begingroup
\makeatletter
\setlength{\@fptop}{0pt}
\makeatother

\begin{figure}[p]
\color{black}
\centering
\includegraphics[width=\linewidth]{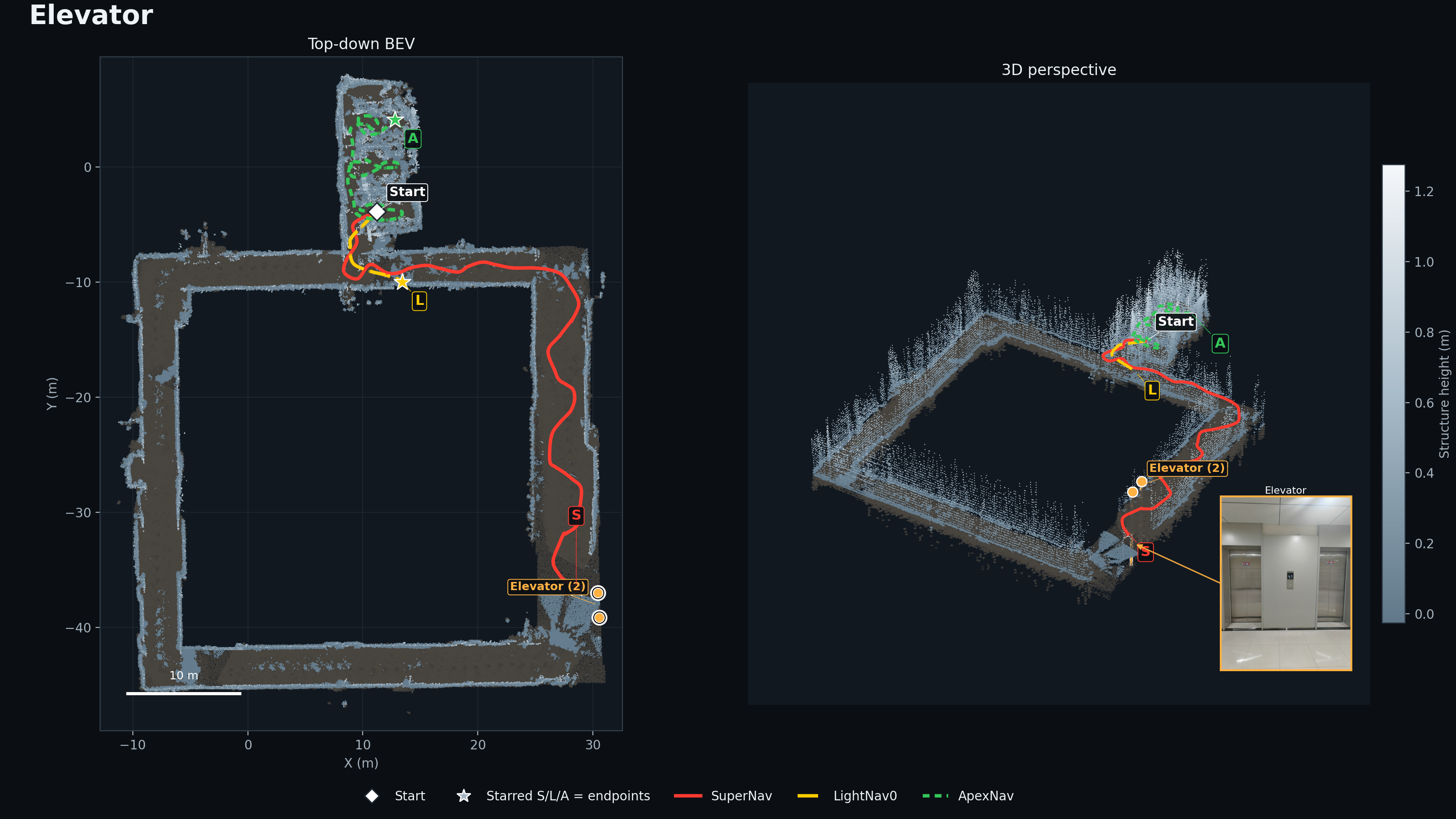}
\caption{Task 1: Find the elevator.}
\label{fig:realworld_tasks_1_2}
\par\medskip
\includegraphics[width=\linewidth]{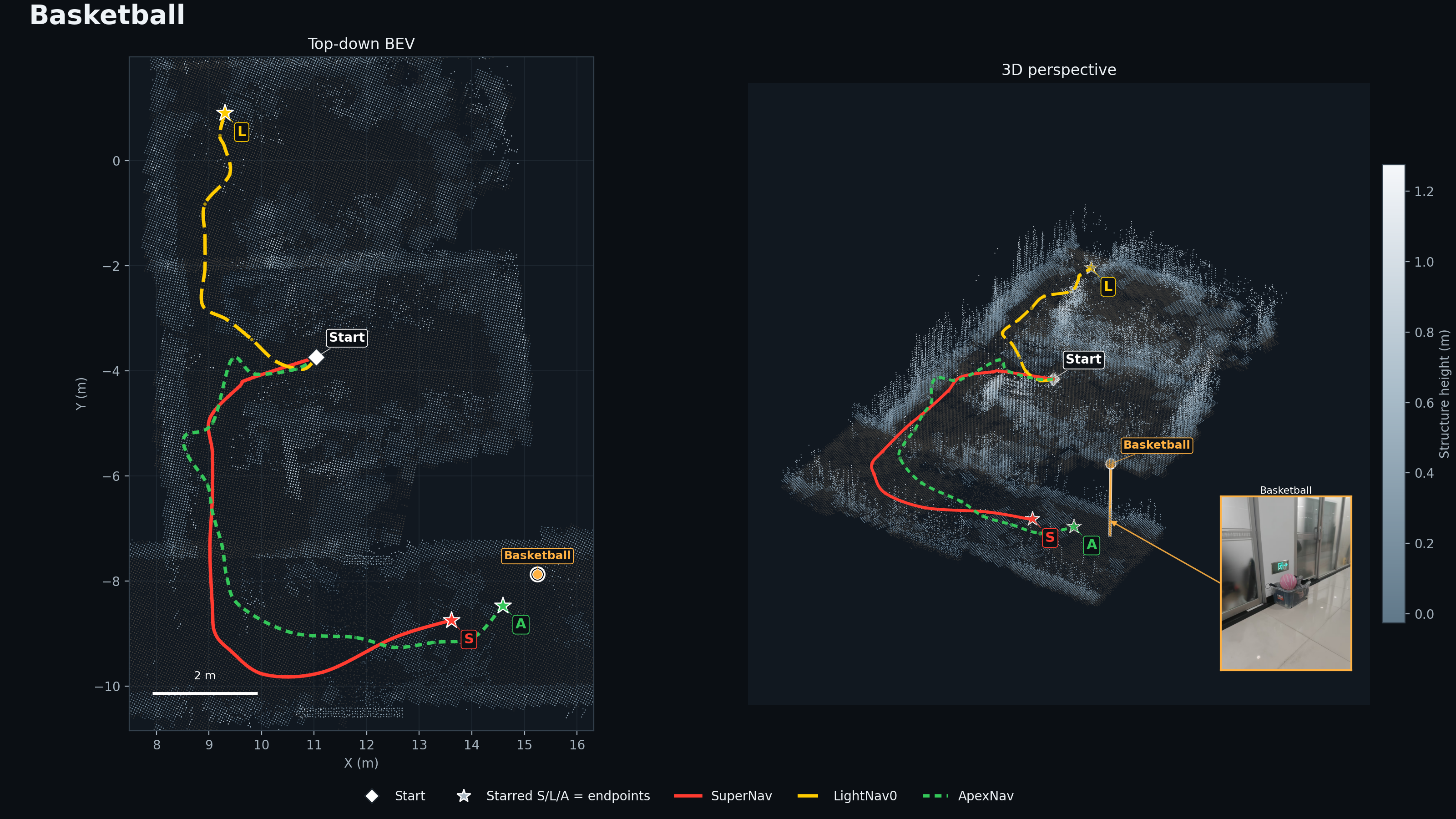}
\caption{Task 2: Find the basketball.}
\end{figure}

\begin{figure}[p]
\color{black}
\centering
\includegraphics[width=\linewidth]{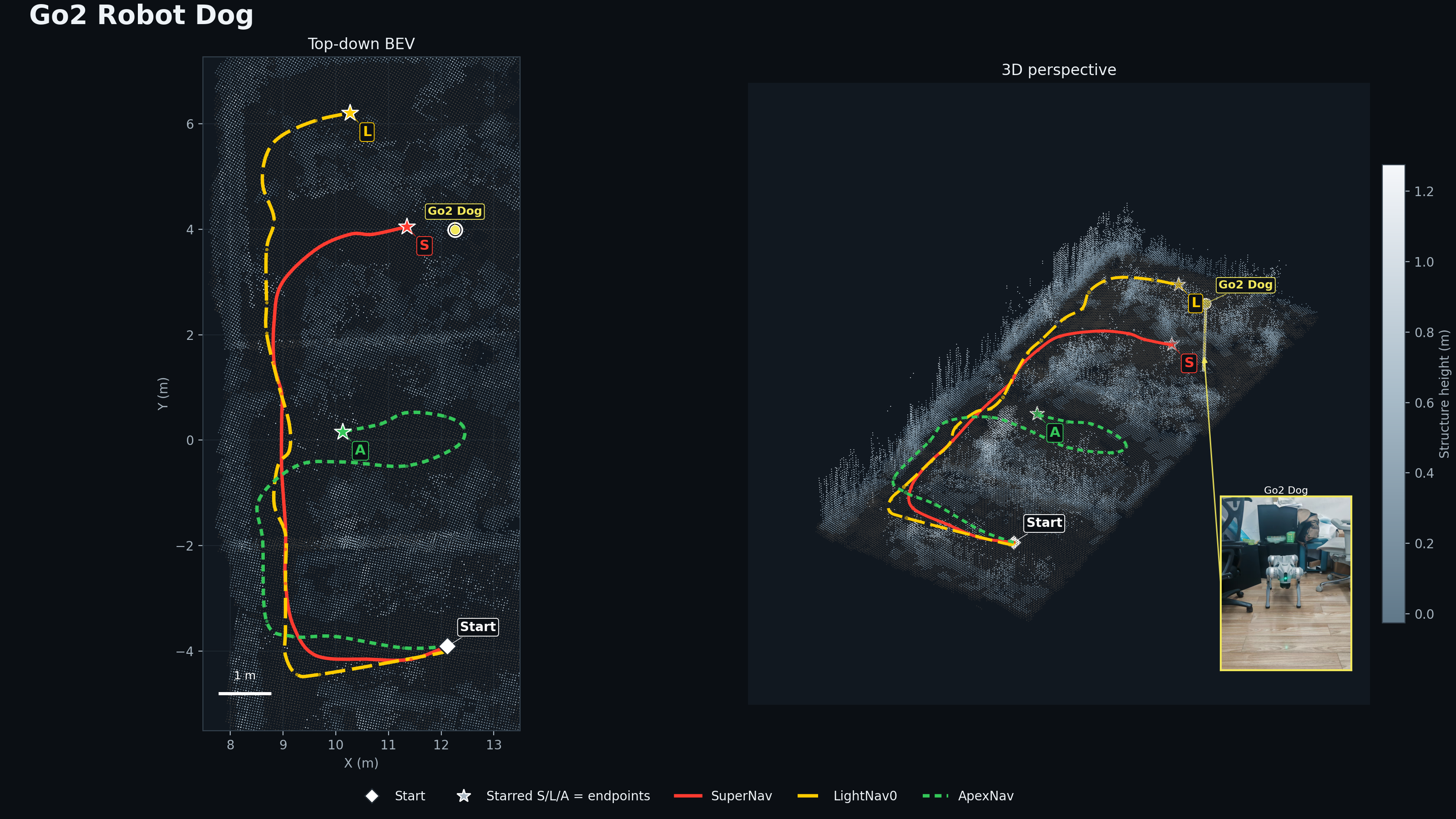}
\caption{Task 3: Find another Go2 dog.}
\par\medskip
\includegraphics[width=\linewidth]{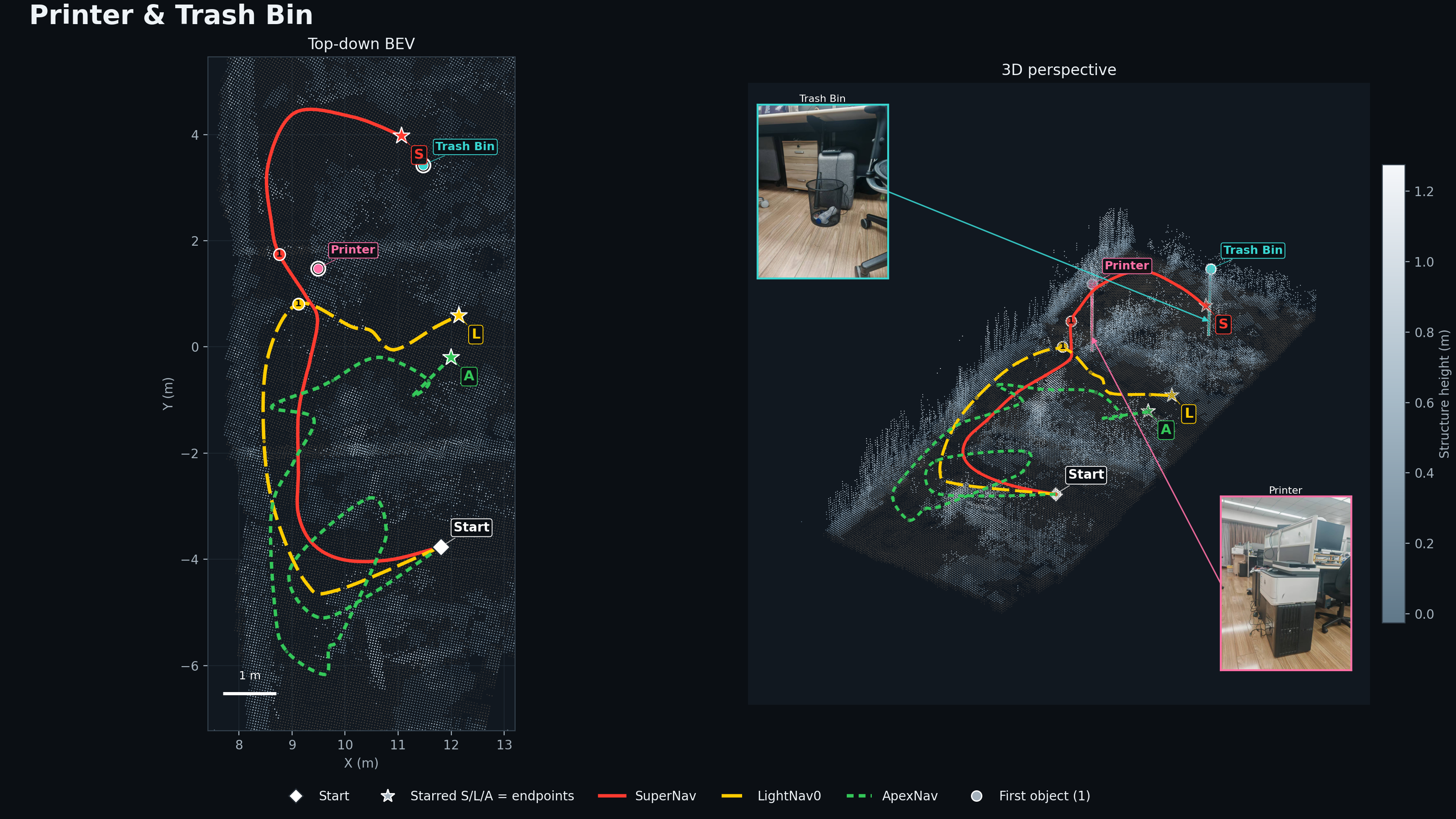}
\caption{Task 4: Find the printer, then the trash bin.}
\end{figure}

\begin{figure}[p]
\color{black}
\centering
\includegraphics[width=\linewidth]{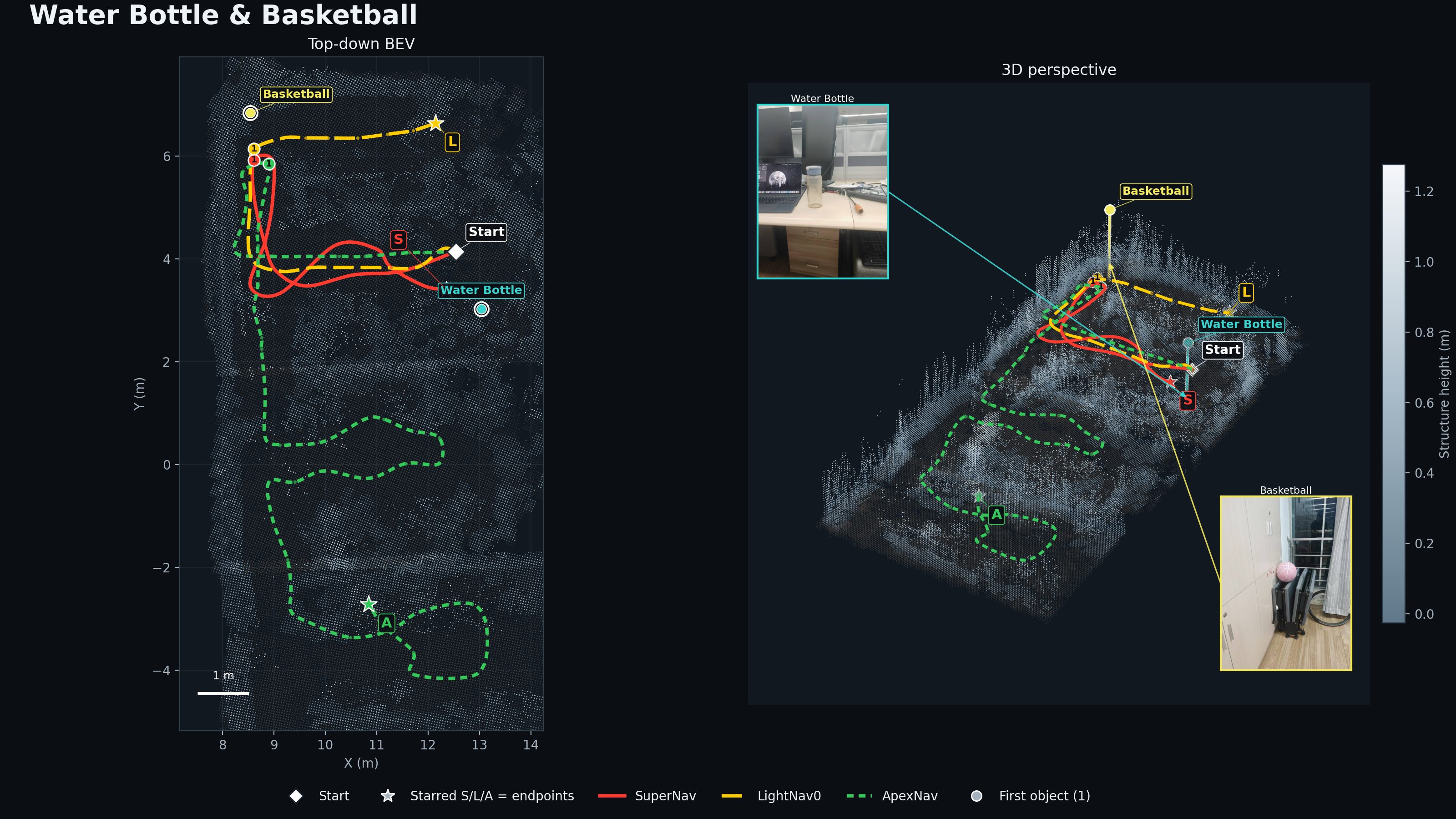}
\caption{Task 5: Find the basketball, then the water bottle.}
\par\medskip
\includegraphics[width=\linewidth]{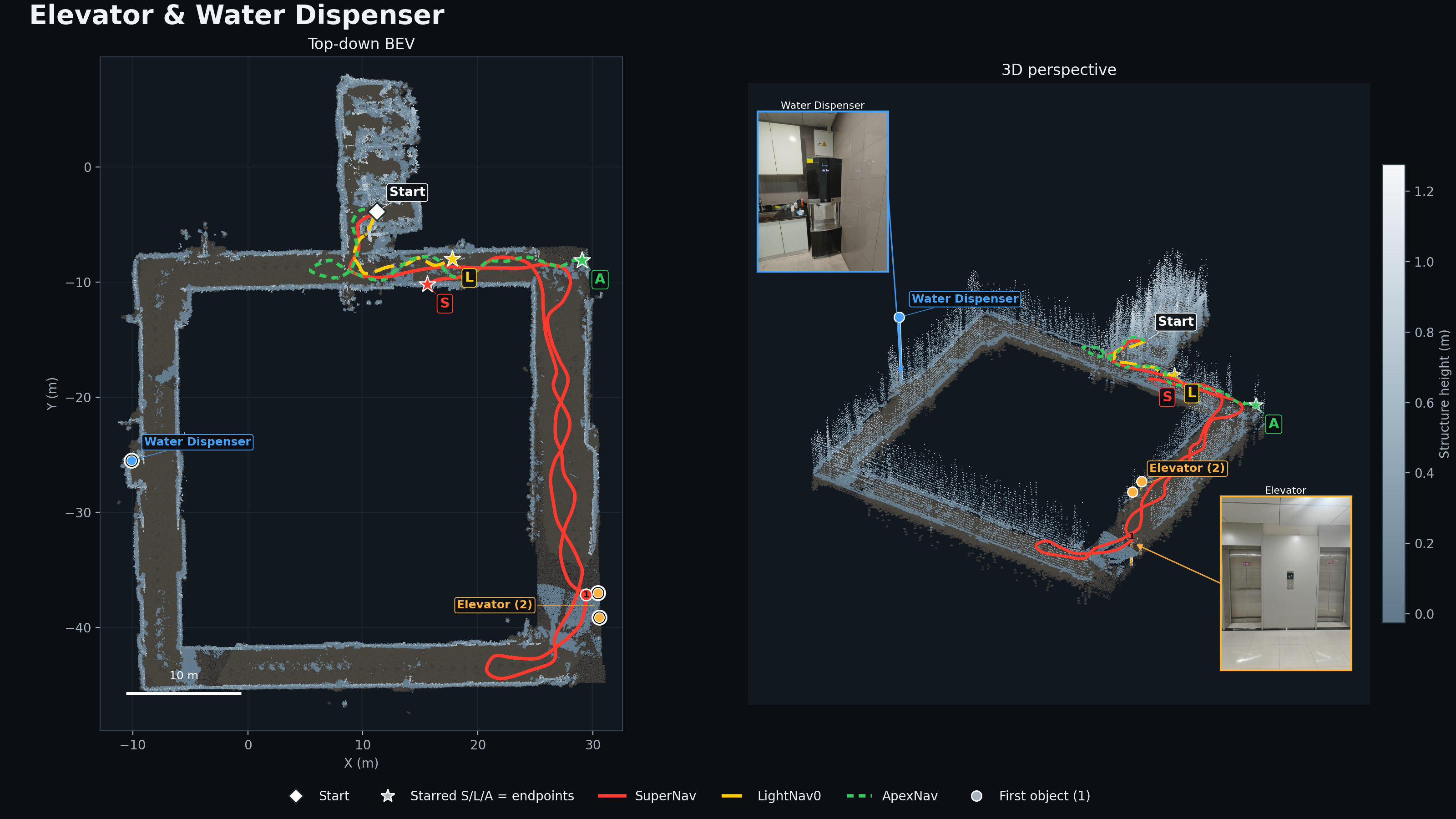}
\caption{Task 6: Find the elevator, then the water dispenser.}
\end{figure}

\begin{figure}[p]
\color{black}
\centering
\includegraphics[width=\linewidth]{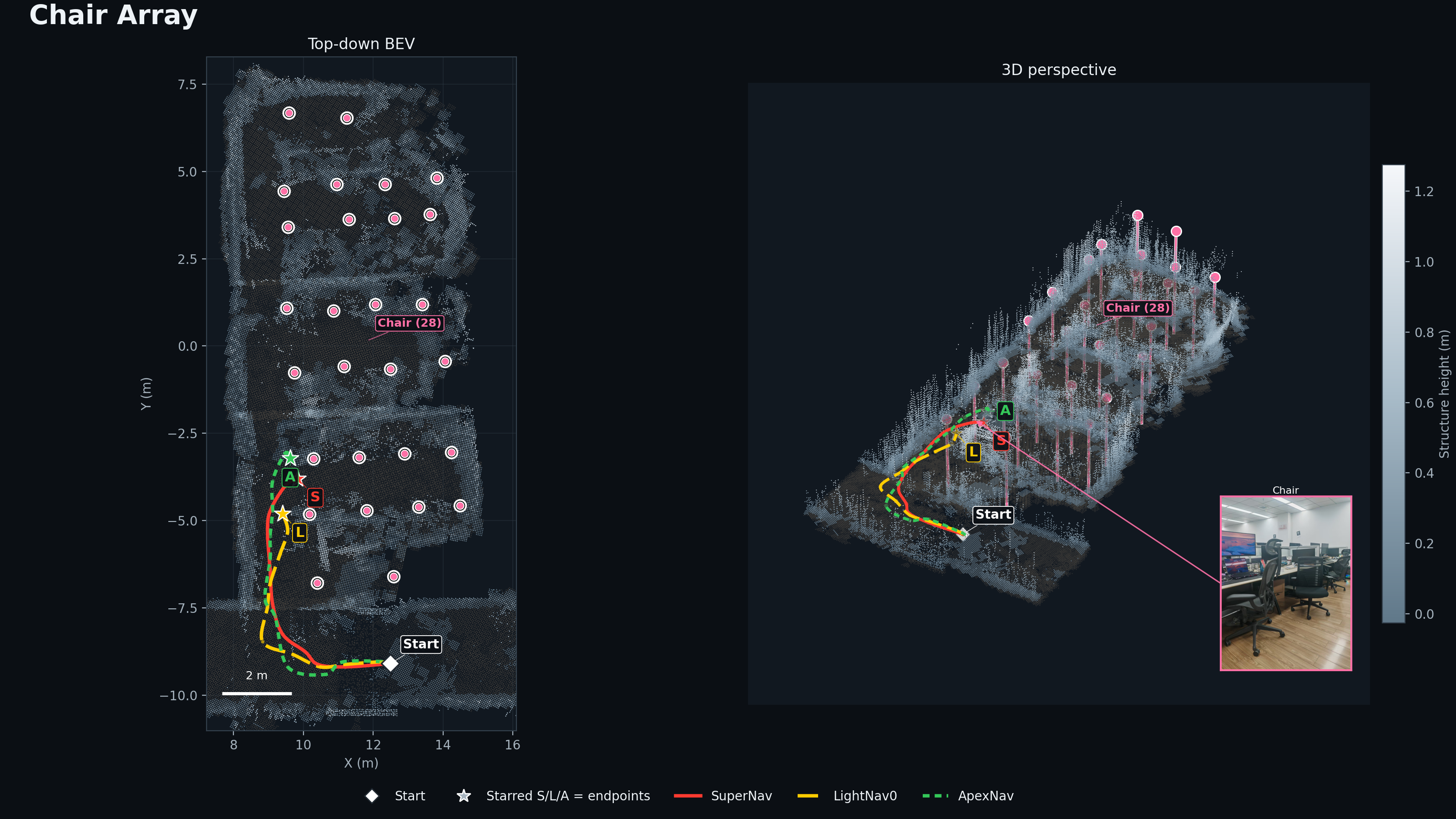}
\caption{Task 7: Find the chair.}
\par\medskip
\includegraphics[width=\linewidth]{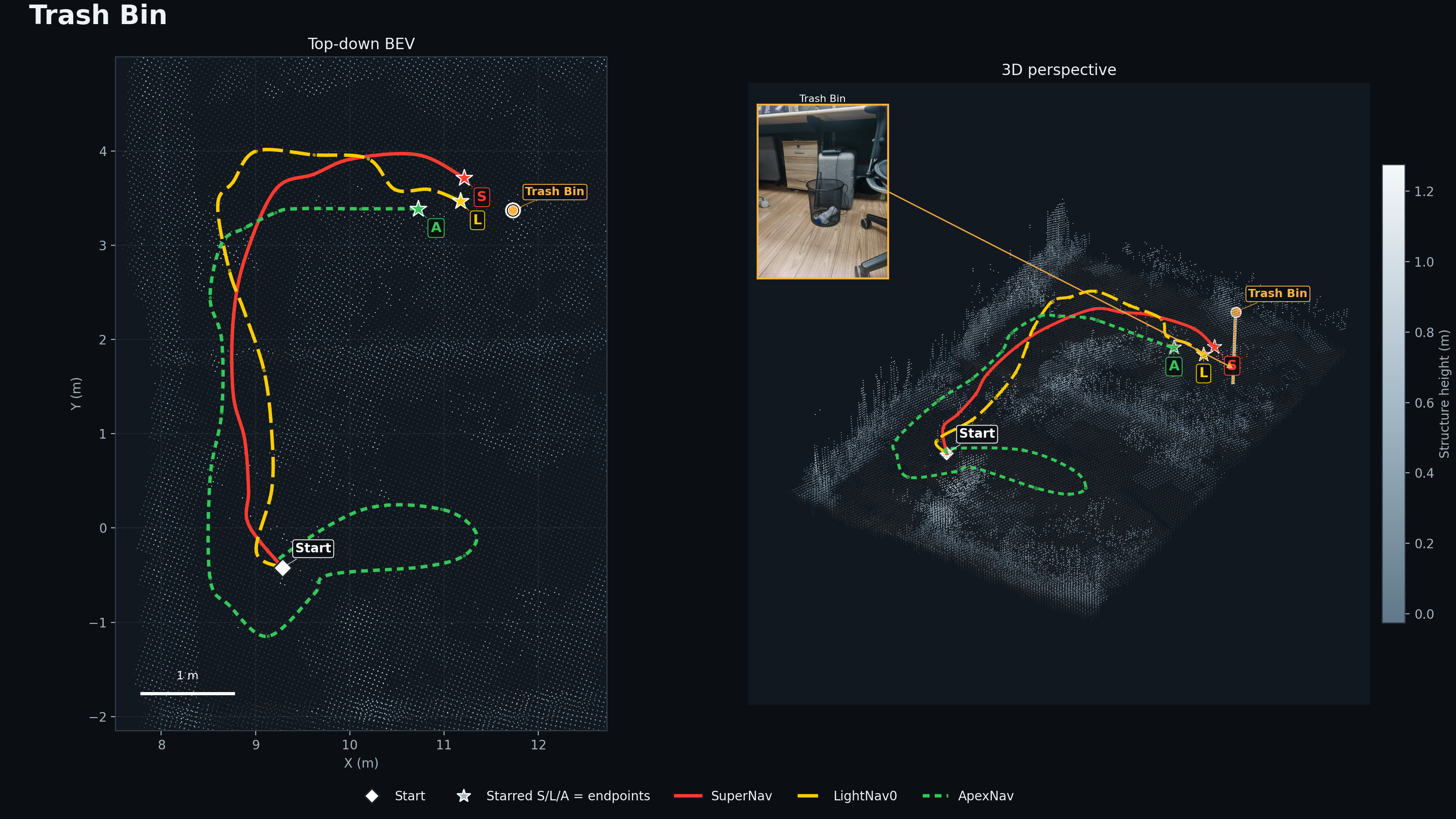}
\caption{Task 8: Find the trash bin.}
\end{figure}

\begin{figure}[p]
\color{black}
\centering
\includegraphics[width=\linewidth]{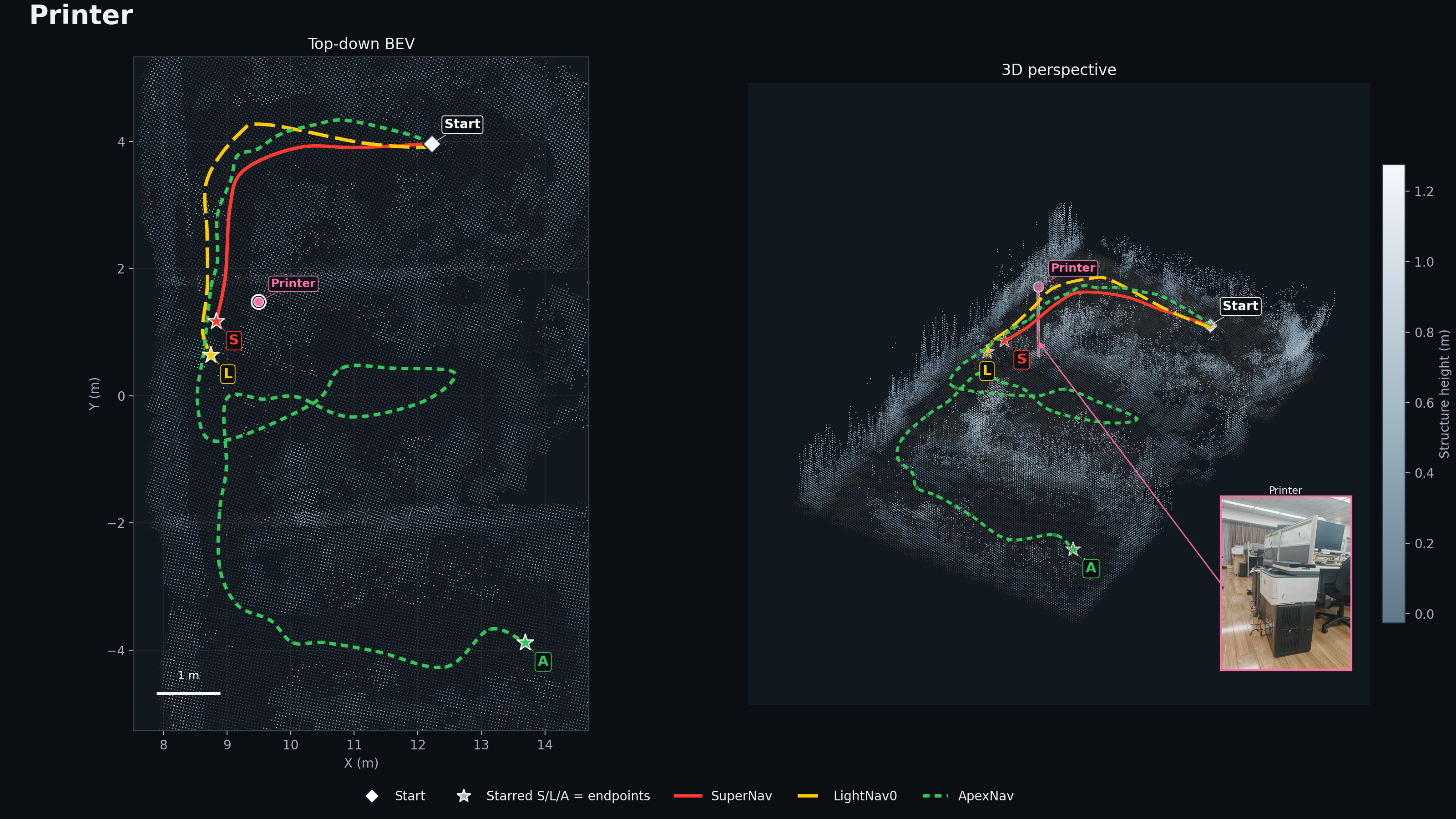}
\caption{Task 9: Find the printer.}
\par\medskip
\includegraphics[width=\linewidth]{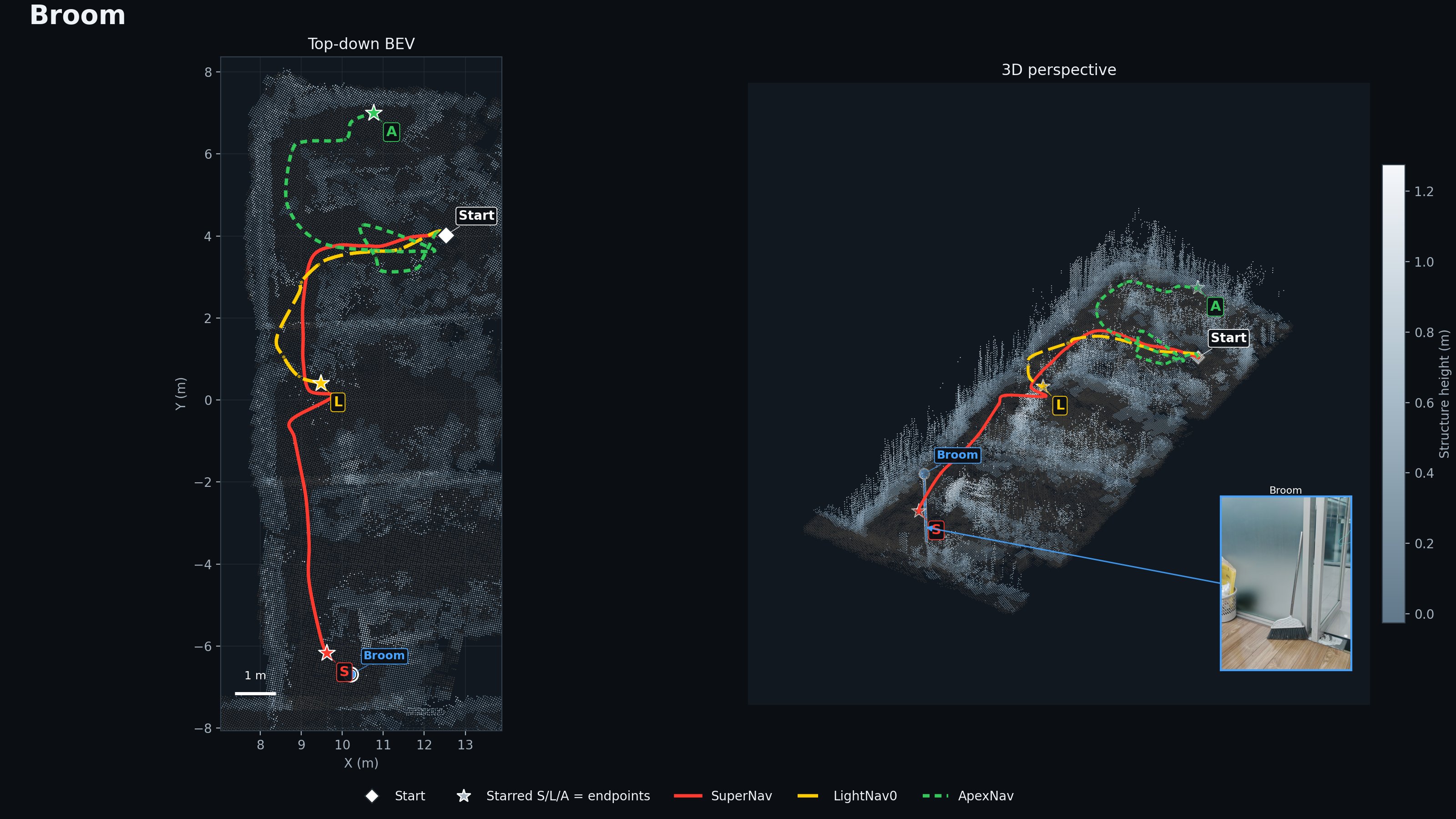}
\caption{Task 10: Find the broom.}
\label{fig:realworld_tasks_9_10}
\end{figure}
\clearpage
\endgroup